\RequirePackage{fix-cm}
\documentclass[twocolumn]{svjour3} 
\smartqed  
\usepackage{graphicx}
\usepackage{subfigure}
\usepackage{amsmath,epsfig,amssymb,graphicx,psfrag}
\usepackage{multirow}
\usepackage{setspace}
\usepackage{tabularx}
\usepackage{rotating}
\usepackage{epstopdf}
\usepackage{placeins}
\usepackage{ifthen}
\usepackage{cancel}
\usepackage[force,almostfull]{textcomp}
\usepackage{lipsum}
\usepackage{stfloats}
\usepackage{paralist}
\usepackage{amsmath}
\usepackage{xfrac}
\usepackage{pdfpages}
\usepackage[multidot]{grffile}  

\usepackage{fancyhdr}
\usepackage{bm}
\usepackage{pgfplots}
\usepackage{pgfplotstable}
\usepackage[greek,english]{babel}

\makeatother

\DeclareSymbolFont{matha}{OML}{txmi}{m}{it}
\DeclareMathSymbol{\varv}{\mathord}{matha}{118}

\usepackage{mathptmx}

\usepackage{acronym}

\usepackage{tikz}
\usetikzlibrary{arrows}
\usepackage{float} 
\usepackage{graphicx}
\usepackage{diagbox}

\usepackage{verbatim}
\newcommand*{\affaddr}[1]{#1} 
\newcommand*{\affmark}[1][*]{\textsuperscript{#1}}
\begin{document}


\newcommand{\myclearpage}{}
\title{ChildBot: Multi-Robot Perception and Interaction with Children}


\author{ Niki Efthymiou\protect\affmark[1,2]  \and Panagiotis P. Filntisis\affmark[1,2]  \and Petros Koutras\affmark[1,2] \and Antigoni Tsiami\affmark[1,2] \and {Jack Hadfield\affmark[1,2]} \and Gerasimos Potamianos\affmark[1,3] \and Petros Maragos\affmark[1,2]
}

\authorrunning{ Niki Efthymiou   \and Panagiotis P. Filntisis$^{1,2}$  \and Petros Koutras$^{1,2}$ \and Antigoni Tsiami$^{1,2}$ \and {Jack Hadfield$^{1,2}$} \and Gerasimos Potamianos$^{1,3}$ \and Petros Maragos$^{1,2}$
}


\institute{
           Niki Efthymiou \at
           Email: nefthymiou@central.ntua.gr
           \and
           Panagiotis P.Filntisis \at
           Email: filby@central.ntua.gr
           \and
           Petros Koutras\at
           Email: pkoutras@cs.ntua.gr
           \and
           Antigoni Tsiami \at
           Email: antsiami@cs.ntua.gr
           \and
           Jack Hadfield \at
           Email: jack103@windowslive.com
           \and
           Gerasimos Potamianos \at
           Email: gpotam@ieee.org
           \and
           Petros Maragos \at
           Email: maragos@cs.ntua.gr
	\and
	\affaddr{\affmark[1]}Athena Research and Innovation Center, Maroussi, 15125, Greece 
	\\
	\affaddr{\affmark[2]}School of ECE, National Technical Univ. of Athens, 15773 Athens, Greece
	\\
	\affaddr{\affmark[3]}Dept. of ECE, University of Thessaly, Volos 38221, Greece \at
	\and
This research work was supported by the EU Horizon 2020 project BabyRobot, under grant 687831.~\cite{babyrobot}
}

\date{ }

\maketitle

\begin{abstract}
In this paper we present an integrated robotic system capable of participating in and performing a wide range of educational and entertainment tasks, in collaboration with one or more children. \textcolor{black}{The system, called ChildBot, features multimodal perception modules and multiple robotic agents that monitor the interaction environment, and can robustly coordinate complex Child-Robot Interaction use-cases. 
In order to validate the effectiveness of the system and its integrated modules, we have conducted multiple experiments with a total of 52 children. Our results show improved perception capabilities in comparison to our earlier works that ChildBot was based on. In addition, we have conducted a preliminary user experience study, employing some educational/entertainment tasks, that yields encouraging results regarding the technical validity of our system and initial insights on the user experience with it.}

\end{abstract}

\section{Introduction}
\label{sec:introduction}

Recently, robotic systems entailing Human-Robot Interaction (HRI) at their core have been gaining momentum not only in research, but also in everyday life. Indeed, such robotic platforms have been entering many different aspects of human lives~\cite{goodrich+2007}, e.g. rehabilitation~\cite{qian2015recent,huo2016lower}, nursing or personal care~\cite{gombolay2018robotic}, education~\cite{pachidis2018social}, and entertainment~\cite{sullivan2017dancing}. \textcolor{black}{In order to achieve a high level of naturalness that resembles human-to-human interaction, robots need to have the ability to perceive and understand the different modalities that people use for communication such as speech or body movements \cite{stiefelhagen2004natural,lucignano2013dialogue}.}

\textcolor{black}{The majority of existing social robotics systems present two major deficiencies: First, they usually incorporate only specific modalities, forcing the users to adapt to the way the system perceives the environment, instead of the opposite. Secondly, they are developed and designed for specific applications and tasks.}

\textcolor{black}{A natural interaction involves the creation of smart adaptive integrated robotic systems capable of multitasking, and with a wide range of perceptual and actuation abilities. An HRI system that would be capable of multitasking, encourages users to design multiform interactive applications that can maintain the interest of the participants undiminished. This is especially important when the participants are children, in the context of education and entertainment (edutainment). In addition, systems leveraging multiple perceptual modalities allow the users to express themselves in their preferred communication means.}

\textcolor{black}{One of the difficulties we have to tackle in order to achieve the above, is the fact that commercial social robots have different capabilities, e.g., the NAO~\cite{NAO} robot is capable and adept in body movements, but incapable of facial expressions, while Furhat~\cite{Furhat} can present a large variety of facial expressions, but can not move outside its installed space. In other cases, robots such as Zeno~\cite{Zeno} are capable of movements and facial expressions, but they are not adept at both (Zeno's body movement lacks in comparison to NAO). Additionally, each social robot has different sensors, constraining the user to specific communication channels.}

\textcolor{black}{Motivated by the above, in this work we present an integrated robotic system that can be used for multiple edutainment applications, called ChildBot. To achieve this versatility, ChildBot incorporates: (i) multiple sensors and perception modules that allow the user to communicate with the robots via multiple channels, and (ii) multiple social robots, leveraging each other's strengths while diminishing their individual weaknesses.}

\begin{figure*}[t!]
\centering
\subfigure{\includegraphics[width=\textwidth]{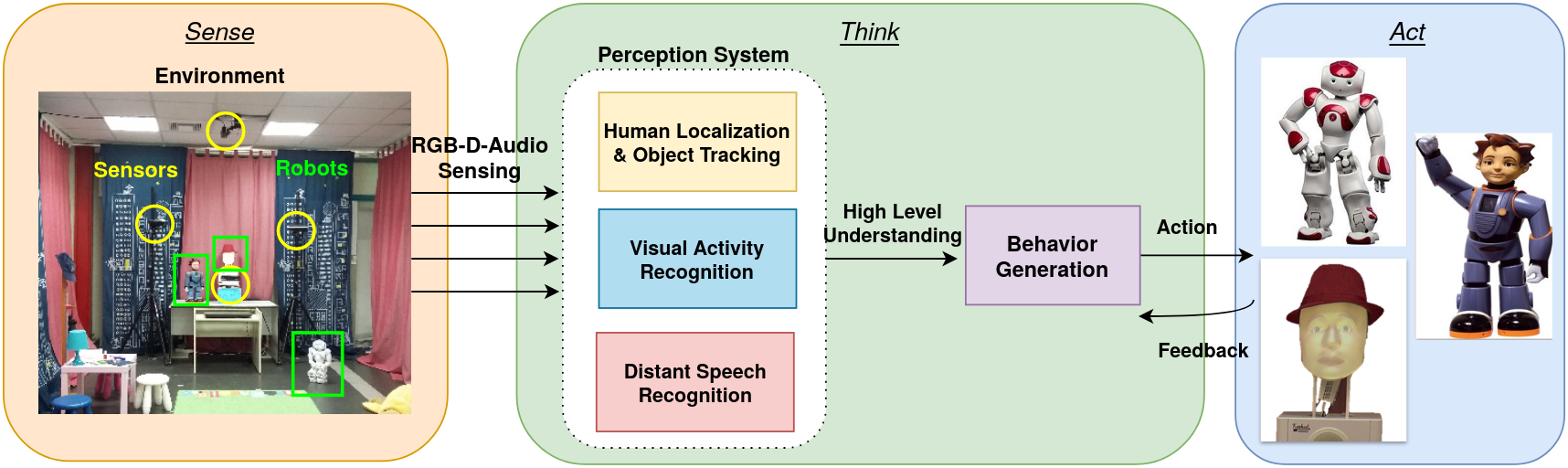}}
\caption{Schematic overview of the ChildBot system during Child-Robot Interaction. The multi-modal information of the child's action is received through a network of sensors placed around an interaction area. The perception system processes it and extracts high-level information about the context of action. Based on this, the behavior generation module decides and controls the robotic agents.}
\label{fig:babysystem}
\end{figure*}

\textcolor{black}{An overview of the proposed system can be seen in Figure \ref{fig:babysystem}. ChildBot is developed using a \textit{Sense-Think-Act} paradigm~\cite{gat1998three} and is indoors based, allowing us to employ external sensors that are arranged throughout a ``smart space" where the interaction takes place. Robot-external sensing can both overcome common HRI problems such as occlusions, and also allows the fusion of different data streams, improving robustness and performance of the different perception modules. This way we also achieve perception of the interaction in a \textit{robot-independent} fashion and bypass limitations of individual robotic systems sensing. In addition, this robot-agnostic architecture can easily accommodate new robots in the loop. The system coordinates a complex and continuous procedure that is required from the moment the child acts until the moment the robot responds and vice versa. Specifically, multi-modal information flows from the sensors (Sense) to the perception modules. Then, high-level information about the context of this action is extracted, and the appropriate response/action is decided (Think). Finally, the system transmits the message to a robotic agent (Act).}

\textcolor{black}{
To showcase the versatility and capabilities of the integrated system, we have designed five different edutainment use-cases. These use-cases are \textit{indicative} and have been designed in order to exploit different components of the system, showcasing the large variety of applications that can be accommodated with ChildBot. The data collected by a pool of 52 children, while playing the aforementioned use-cases with the robots, allow us to objectively evaluate the performance of each module of the ChildBot system regarding its perception capabilities (accuracy-wise). Furthermore, our initial evaluation on the user experience shows encouraging results towards a complete well-designed subjective evaluation in the future.}

\textcolor{black}{ChildBot is an improved integrated extension of a set of preliminary conference publications by the authors on specific problems of multi-robot perception and interaction, and it presents a wide-application Child-Robot Interaction (CRI) system able to manage multitasking interaction autonomously and envelop a plethora of purposes, such as edutainment~\cite{Tsiami2018Multi,Tsiami2018FarField,Efthymiou2018,Hadfield2018}. The work presented here has integrated the previous works under a single and modular three-layer multi-robot architecture~\cite{gat1998three}, includes improved perception modules, and is evaluated extensively on a larger corpus that contains spontaneous children data, more representative of CRI.
To summarize, we highlight the most important contributions of the presented work:}

\begin{itemize}
\item \textcolor{black}{\textit{An integrated system for HRI} has been designed and implemented by leveraging multiple robotic agents. The modular three-layer system architecture integrates multiple sensors, numerous perception modules and different robotic agents, and culminates into a multi-application autonomous HRI system.}

\textcolor{black}{\item \textit{Perception modules for multimodal scene understanding} have been developed and adjusted in specific CRI conditions by incorporating novel approaches and extensive studies. Audio-visual active speaker localization, 6-DoF object tracking, visual activity recognition, and distant speech recognition are necessary for analyzing and tracking human behavior over time in the context of their surroundings. The perception modules of this system have been developed according to and sometimes exceed the state-of-the-art of the constituent technologies, as shown by our objective evaluations.}

\textcolor{black}{\item \textit{Spontaneous children data during CRI} have been collected and used for system evaluation. Indicative use-cases have been defined and implemented in order to showcase the large range of applications that ChildBot can be used for.} \textcolor{black}{The data collected have allowed for an extensive objective evaluation of the ChildBot capabilities during real use-case scenarios, as well as a preliminary user experience study with promising results.}

\end{itemize}


\section{Related work}

Many research projects have aimed at developing robots both in ambient assisted living environments~\cite{mayer2012examples}, as well as in well-defined and constrained environments e.g., in bathrooms for assistive bathing~\cite{zlat+17}. Some robotic agents act as companions to improve quality of life, assist with mobility, or complete household tasks~\cite{FISCHINGER201660,nani2010mobiserv}, while others are designed to help people live independently and serve themselves when they face difficulties due to disabilities or old age~\cite{kardaris2016platform,dementia}. Nevertheless, the intervention of robots in human life remains a controversial issue~\cite{shishehgar2018systematic,Frennert2014,ROBINSON2013661,wu2012designing}.  

Regarding educational CRI applications, many previous works focus on the theoretical exploration of different social robot behaviors in the learning experience, without delving deeply into the technical aspects, but mostly using off-the-shelf solutions for environment perception. An immediate result of this is the fact that the interaction space is constrained. In~\cite{kennedy2015higher}, a study took place that involved children playing an educative mathematics scenario with a NAO robot. In~\cite{saerbeck2010expressive}, Saerbeck et al. studied the effect of social robot behavior on the learning performance of the subjects, in the context of a language learning task. Similar studies can be found in~\cite{gordon2015can}, while in~\cite{robins2005robotic} a humanoid robot was employed to interact with autistic children.

Notable works that have also focused on the robot perception aspect of CRI include the ALIZ-E project~\cite{belpaeme2012multimodal}, where a complete framework was developed for multimodal CRI, and the NAOTherapist platform~\cite{pulido2017evaluating} for upper-limb rehabilitation sessions for children with physical impairments. \textcolor{black}{A similar platform was built in the INSIDE project~\cite{inside}, where a multimodal perception system was developed to allow autonomous interaction of a NAO robot with ASD children. Other similar projects include EMOTE~\cite{emote}, where the perception system focused mostly on visual communication, L2TOR~\cite{l2tor}, where a NAO robot capable of multimodal perception assumed the role of a second-language tutor, and the EASEL~\cite{easel} educational CRI project.} Esteban et al.~\cite{esteban2017build} built a multi-sensor system for autonomous interaction of a NAO robot with autistic children to perceive different features during an interaction such as gaze estimation, action recognition, and object tracking. The capabilities of the system were sufficient for the presented tasks, but limited for a more generic interaction, and the system lacked in real use-case evaluation. In~\cite{marinoiu20183d}, Marinoiu et al. introduced an action and emotion recognition system by exploiting 2D and 3D pose estimation methods and evaluated it in a large-scale dataset of robot assisted therapy sessions of children with autism. \textcolor{black}{The ANIMATAS~\cite{animatas} project is also worth mentioning, focusing on training researchers to advance human-machine interaction.}

A general review of the perception methods used for HRI in social robots until 2014 is presented in~\cite{yan2014survey}. Three important issues associated with perception systems are highlighted there: the need for developing perception systems in real environments with real data, the requirement of creating good representations in accordance with the context of the interaction, and the demand of combining an efficient perception system with reasonable robot responses in order to create pleasant HRI experiences. Some works that attempt to develop perception systems for HRI are Zaraki et al.~\cite{zaraki2017design}, where low-level and high-level features are combined to detect a range of human-relevant features that appear during a real use-case procedure, and Valipour et al.~\cite{valipour2017incremental}, where a  novel  paradigm for  incrementally improving visual perception of a robot during an HRI experience is proposed.

\textcolor{black}{
Aiming to increase performance, flexibility, and robustness, ChildBot consists of multiple robots and multimodal perception modules designed for and adapted to children, and thus allows interaction in a relatively large space for a variety of edutainment tasks.}
\textcolor{black}{Parts of the ChildBot system have been based on our preliminary previous works, where an early design and development was presented. More specifically, a preliminary setup and evaluation of a basic architecture in a few use-cases has been presented in~\cite{Tsiami2018Multi}, while~\cite{Tsiami2018FarField} has focused mainly on multi-party interaction via speech.
In~\cite{Efthymiou2018} the techniques for multi-view fusion of action recognition have been explored more in depth, and finally, \cite{Hadfield2018} has focused on the development of tracking algorithms, essential in interactive tasks between a child and a robot.
In all these previous works, a limited and specific functionality of the system has been investigated, and evaluation has been carried out employing data acquired in a strictly controlled annotation procedure, and not spontaneous data from real interaction.}

\textcolor{black}{The current paper integrates all these different modules from previous works under the same, unified system, using a three-layer architecture. The modules can now work both in isolation and in synergy, and due to the system modularity, the addition or removal of a component is an easy task. Moreover, apart from the integration, a lot of effort has been dedicated to improving performance of the perception modules. For this reason, more sensors have been included, and ablation studies have been carried out, in order to validate the plausibility of the employed modules. Most importantly, contrary to all other previous works, the modules evaluation has been performed employing real-time spontaneous children data (see Sec.~\ref{section:data}), which are more challenging.}


\myclearpage

\section{System Overview}
\subsection{Perception System}

\begin{figure*}[h!]
\centering
\subfigure{\includegraphics[width=\textwidth]{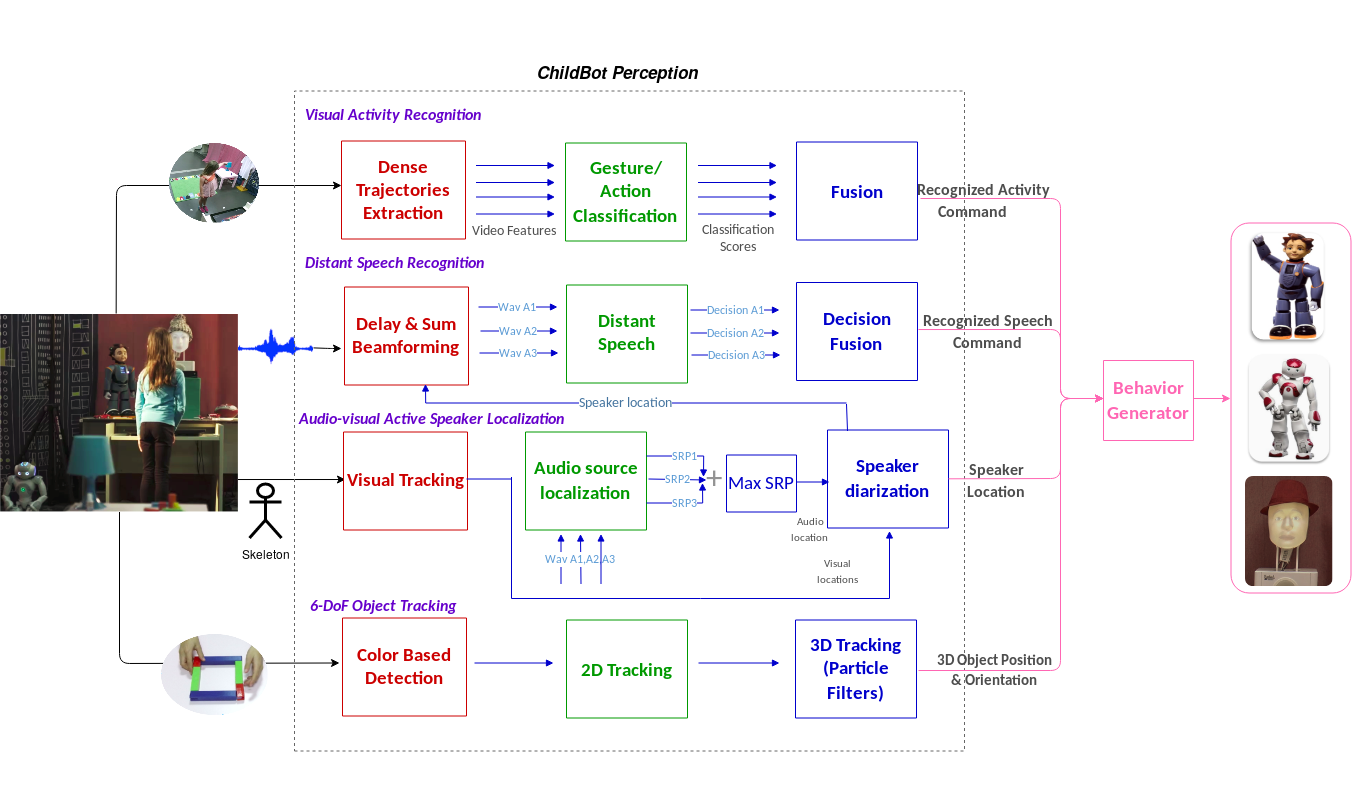}}
\caption{Overview of ChildBot perception modules including \textit{Audio-Visual Active Speaker Localization and 6-DoF Object Tracking}, \textit{Visual Activity Recognition}, and \textit{Distant Speech Recognition}. ``A'' refers to microphone array and ``SRP'' to Steered Response Power. The modules are employed during CRI to monitor the multiple aspects of human behavior and then their outputs are fed to the robotic behavior generator module.}
\label{fig:action}
\end{figure*}

 This section focuses on the ChildBot perception system which provides a global and effective supervision of a progressing CRI. A core audio-visual processing technology acts as eyes, ears, and brain for the robots in use by incorporating different recognition and tracking modules. By analyzing and tracking human behavior in the context of a structured CRI experience, we target to establish common ground and share goals with the children. 
 
 The main overview of the robot-agnostic perception system can be seen in Figure \ref{fig:action}, consisting of three main modules: \textit{Audio-Visual Active Speaker Localization and 6-DoF Object Tracking}, \textit{Visual Activity Recognition}, and \textit{Distant Speech Recognition}. Four Kinect V2 sensors capture a detailed raw data representation of the environment and feed it into the perception system. The Kinect V2 sensors have been placed at different positions and viewing angles, in order to sufficiently cover the entire environment, tackle occlusion problems (self or from objects), as well as offer multiple viewpoints for visual perception. The raw data that are recorded from the sensors include both RGB and Depth, as well as audio from the microphone array (4 microphones) of each Kinect. The spatial arrangement of the sensors is presented in Figure \ref{fig:1a}.
Subsequently, we present an overview of each perception module:

\textbf{Audio-Visual Active Speaker Localization and 6-DoF Object Tracking:} In order to have a natural interaction between robots and humans, robotic awareness of active speaker localization, as well as important object detection and tracking are essential. An effective audio-visual method for active speaker localization in HRI scenes has been developed to track the children's body, by leveraging audio information in addition to visual information. Moreover, a module for object recognition that can detect multiple toys based on their colors and size of color regions has been incorporated in the system. Finally, a 3D tracking method has been designed for providing both the 3D location and the orientation of rigid objects.

\textbf{Visual Activity Recognition:} A visual frontend has been developed for recognizing hand gestures that accompany everyday communication, as well as more general body movements that convey specific meanings. The multiview visual activity recognition module is able to successfully recognize the child's activity, while wandering around the room and interacting with the robots and objects. The gesture recognition version of the module aims at identifying hand gestures that deliver a conceptual message during the interaction, such as waving at the robot hello or asking the robot to come closer. On the other hand, the action recognition version targets child body movements that form complex meanings, such as pantomimic movements.

\textbf{Distant Speech Recognition:} A multisensory distant speech recognition (DSR) system in Greek has been developed to enable CRI via speech. As close-talking microphones are not convenient for children and restrict their movements, we take advantage of the multiple microphone arrays located around the room recording audio, while at the same time the children can move freely and communicate hands-free with the robots. In order to make the DSR more robust and exploit the distributed microphone arrays we experiment with adaptation and fusion. 

The high-level understanding that is obtained by the perception modules is then fed to the Dialog Manager, along with extra input from a Touch Screen, which is used during the interaction as an extra means of communication. According to its input, the Behavioral Generator then decides on the action that the multi-robot system should do, and forwards its decision to the actuators. The actuators in their turn respond with information back to the system.

In order to create a detailed picture of the ChildBot system, we proceed by describing extensively each perception module. Following this, we present the system architecture, the intercommunications, and the dialog management module in order to describe our complete system for CRI. 

\subsection{Perception Modules}
\label{ss:perception}
\subsubsection*{Audio-Visual Active Speaker Localization}

When analyzing and understanding an auditory or audio-visual scene that consists of multiple speakers, sound and speaker localization are necessary for tracking. In addition, the speaker's location is required for beamforming and for guiding the robot's attention/head in a multi-party scenario, in order to achieve a natural and intuitive interaction. Although visual tracking can be more precise, it does not suffice when the active speaker/speakers have to be localized among other non-speaking persons in an audio-visual scene.

Various techniques for audio speaker localization~\cite{anguera12} have been proposed in literature. Some of them have been specifically adapted to HRI setups~\cite{evers17source,cech2013active} for microphones mounted on robots. In our multi-robot case, microphones are external to the robots, and a fast algorithm is needed due to the real-time nature of our system. 
Thus, a real-time 3D audio localization SRP-PHAT (Steered Response Power - Phase Transform) system based on~\cite{DoSiYu07,BrOmSv07} has been developed,
which is robust to noise and errors. 
Regarding audio-visual speaker localization~\cite{gebru2017,garau2010,minotto2015}, several methods have been developed for RGB cameras, most of them employing Bayesian filtering techniques or fusion between audio and video features. %
In our case, visual tracking is accomplished via skeleton tracking, developed for Kinect sensors.

\begin{figure*}[!tb]
\centering
\subfigure[Experimental setup\label{fig:1a}]{\includegraphics[width=0.35\linewidth]{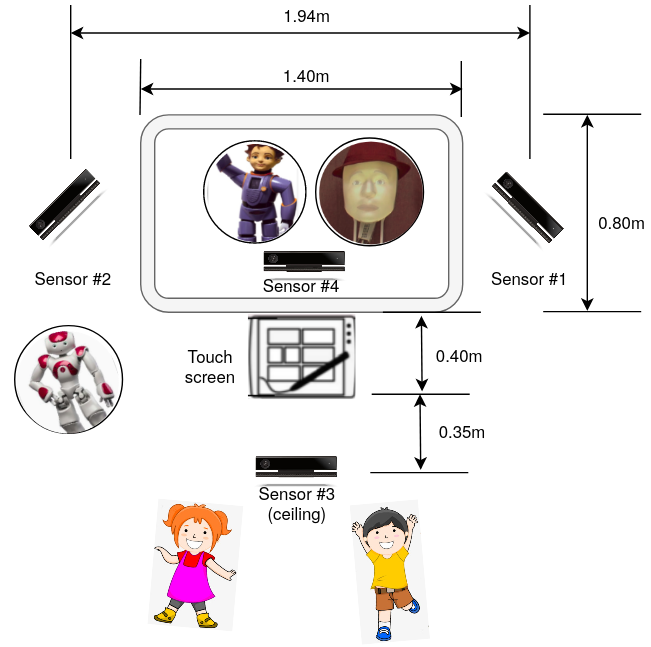}} 
\subfigure[Audio source localization\label{fig:2a}]{\includegraphics[width=0.21\linewidth]{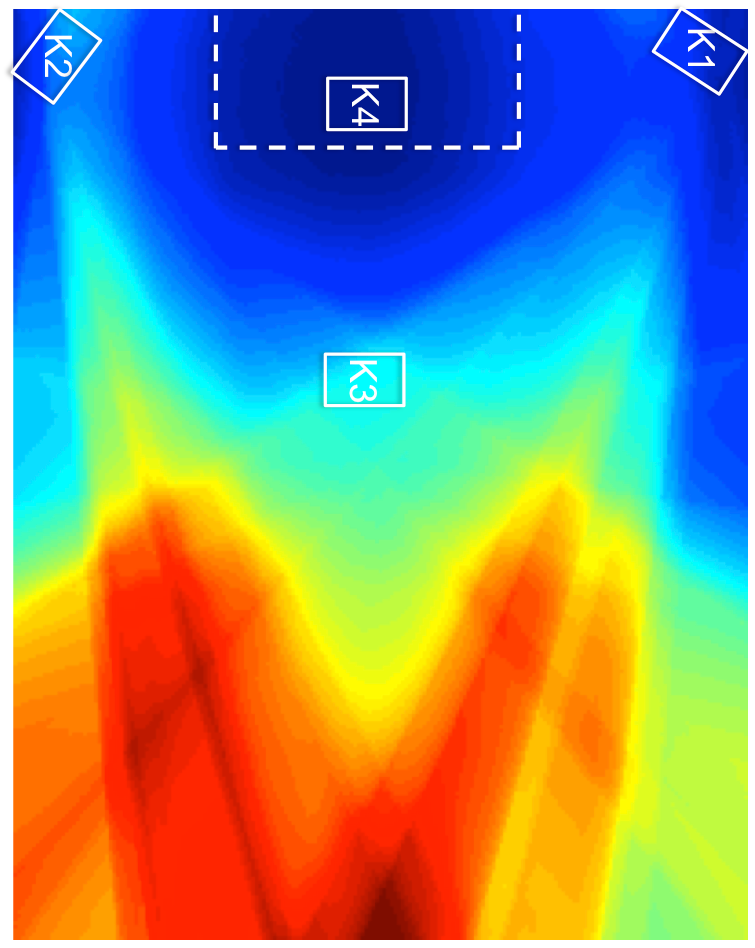}}
\subfigure[Visual person localization\label{fig:3a}]{\includegraphics[width=0.21\linewidth]{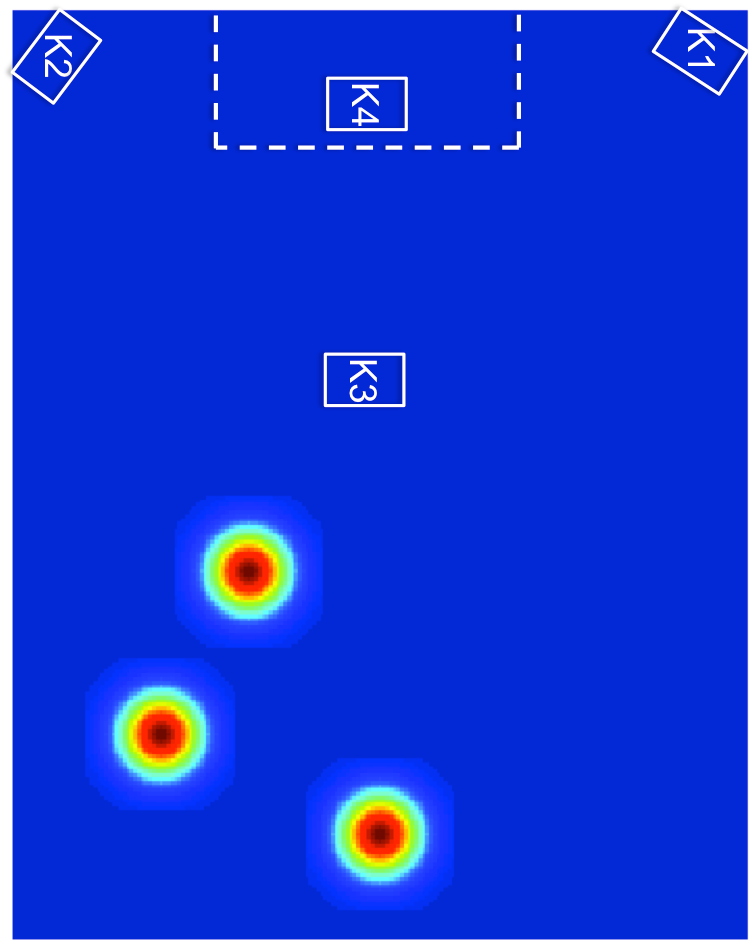}}
\subfigure[Audio-visual active speaker localization\label{fig:4a}]{\includegraphics[width=0.21\linewidth]{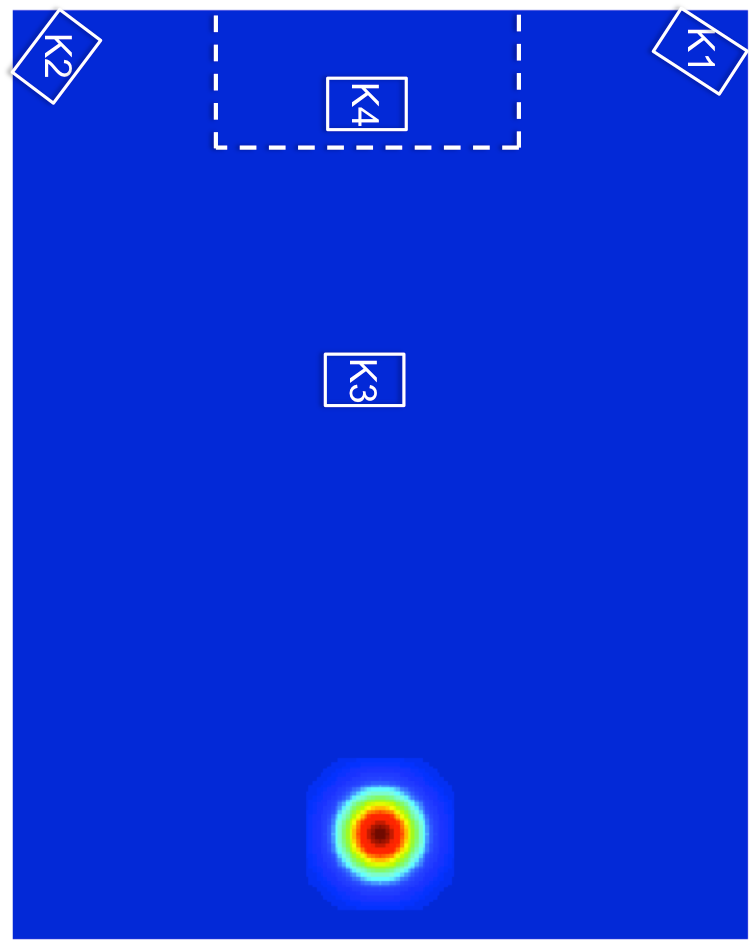}}
\caption{(a) Spatial arrangement of four sensors. (b-d) An example of audio-visual active speaker localization. The SRP output is shown with high values in red. Positions of the table and the four Kinects are also shown.}
\label{fig:slocexample}
\end{figure*}

Our audio-visual active speaker localization that exploits the 3D skeleton and the microphone arrays is performed as follows: 
Person tracking is first achieved by retrieving the 3D skeletons from all persons present in the audio-visual scene. Auditory source localization via SRP-PHAT provides information concerning the speakers. The final active speaker localization is performed by choosing the visual locations that are closest to the auditory ones. 
The speakers positions are then used 
for the robot's attention guiding by turning the robot's head towards the active speaker. An example of audio-visual active speaker localization can be seen in Figure \ref{fig:slocexample}.

\subsubsection*{6-DoF Object Tracking}

\begin{figure}[b!]
	\centering
	\includegraphics[width=0.95\linewidth]{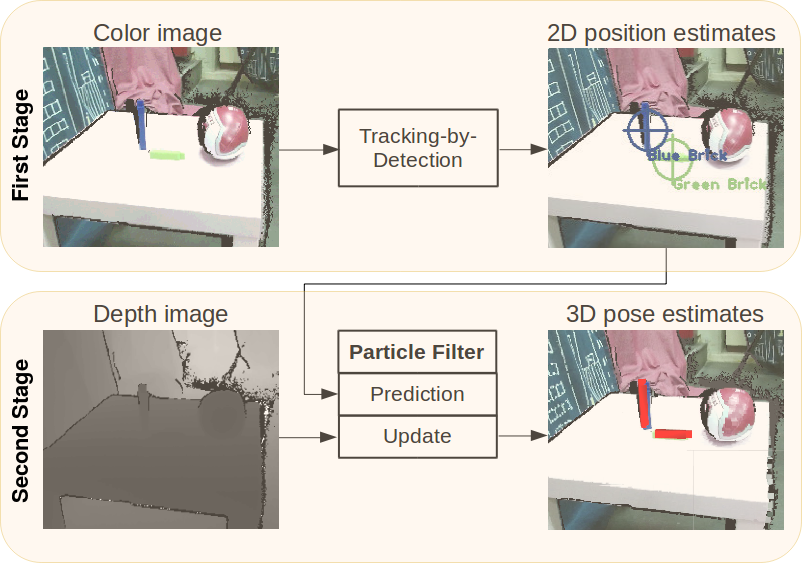}
	\caption{Overview of the implemented 6-DoF object tracking module \textcolor{blue}{(the bricks are tracked).}}
	\label{fig:ObjectTrackingOverview}
\end{figure}

In certain cases, children and robots may be expected to interact with various movable objects. Thus, aside from human localization which has been described above, the robot must possess an understanding of the configuration of these objects. We have developed a method for robustly tracking the 6-DoF poses of multiple objects in real-time. The main idea is to crudely detect the objects in a computationally cheap manner, and then use the detected positions to infer each object's 3D pose. 
The used objects are known beforehand, meaning that their shape and appearance models are predefined. The developed tracker consists of two stages: the first involves a tracking-by-detection scheme upon the color stream to locate the objects on the image plane, while the second performs an operation on the depth data to refine the first stage output and infer the remaining variables related to the object rotations. The basic architecture is presented in Figure \ref{fig:ObjectTrackingOverview}.

During the first stage, our approach uses a simple color histogram model to detect object regions, though depending on the object characteristics a variety of features could potentially be used. The histogram models are defined offline and remain unchanged during the entire tracking. The hue and saturation of the HSV color space were used to introduce sufficient robustness to brightness changes. Assuming the histograms are normalized, they define a probability distribution over the color space. Therefore, a probability map can be generated over the latest color image, which after thresholding and morphological filtering leads to a binary mask that contains the most likely object regions in the image. We choose to retain the region with the largest area, under the assumption that the remaining regions will correspond to noisy artifacts or irrelevant background objects. 
The center of the chosen region is taken as the object location, and a confidence score $s_k$ is produced for object $k$ .

Once the object locations have been detected on the image plane, the tracker's second stage consists of estimating the 6-DoF poses with the help of the newest depth image. The developed tracker employs particle filters and is closely based on the algorithm proposed in \cite{Wut+13}, where the hidden states are augmented with a set of binary variables that model the occlusions at each pixel. 
Using the camera inverse perspective mapping and the depth image, \textcolor{black}{we transform the $k$-th object's 2D position estimates $\textbf{p}_k$ into 3D estimates $\textbf{P}_k$. The input vector for each object $k$ is then $\textbf{u}_k =( \textbf{P}_k - \textbf{r}_k) \cdot s_k$, where $\textbf{r}_k$ is the particle's position estimate from the previous time step, and $s_k$ is the confidence score produced by the tracking-by-detection module.} Using a Rao-Blackwellisation technique~\cite{murphy2001rao}, only the pose variables need to be sampled, while the occlusion variables can be marginalized out analytically. \textcolor{black}{ 
In order to prevent collisions in the object estimated configuration, the observation model is weighted by a factor that depends on the existence of mesh intersections in the particle estimates. If no intersections exist, this factor is set to $1$, otherwise it is set to $0.01$.}

\subsubsection*{Visual Activity Recognition}

For understanding nonverbal communication, an efficient multi-sensor visual activity recognition frontend has been developed by experimenting with Dense Trajectories features \cite{wang_action_2011} along with different encoding methods and fusion schemes for visual information processing. Dense Trajectories have been chosen over convolutional neural network pretrained features, as the actions included in the state-of-the-art databases are not similar to those of children and the fine-tuning of pretrained networks does not perform adequately, since in our case we have limited data from real-world CRI~\cite{Efthymiou2018}. 

\begin{figure*}[!tb]
\centering
\includegraphics[width=1\textwidth,]{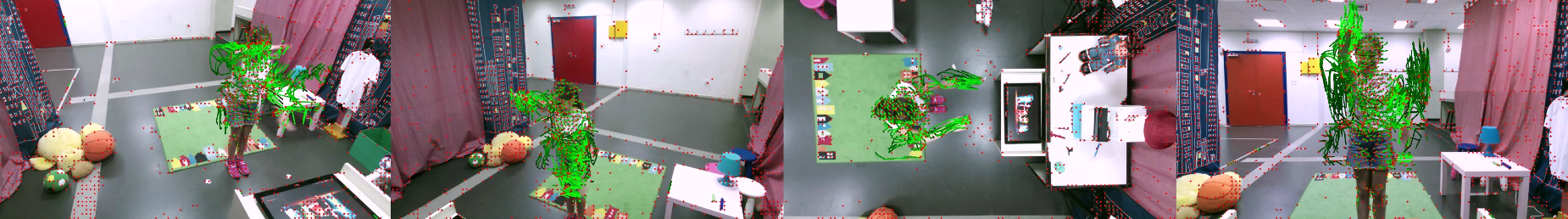}
\caption{Example of the extracted Dense Trajectories from different sensor perspectives while the child is performing the swimming pantomime (see Sec.~\ref{section:usecases}).}
\label{fig:dense}
\end{figure*}

The main goal of this investigation is to establish a robust framework for tackling different tasks, such as generic body movements performed by kids, with limited training data. We have implemented two different versions of the module in the ChildBot system that work independently, one for gesture recognition and one for action recognition. Although the pipeline for both versions is the same, they are trained, tested, and enabled separately for hand gestures and more general body movements respectively. An example of the extracted Dense Trajectories during a pantomime performance (see Sec.~\ref{section:usecases}) is presented in Figure \ref{fig:dense}.

In a more detailed view of the system, the recorded RGB information from each of the four RGB cameras is sampled frame by frame for the various children visual actions. Feature points are sampled for each frame on a grid and are tracked through time based on dense optical flow~\cite{farneback2003two}. Multiple spatial scales are used for the sampling and the tracking independently, while the trajectories are pruned to a fixed length to avoid drifting. The computed features include the Histograms of Optical Flow (HOF)~\cite{laptev_learning_2008} and the Motion Boundary Histograms (MBH)~\cite{wang_action_2011} on both axes (MBHx, MBHy). 

Afterwards, the features are encoded employing either the zero-order statistics Bag-of-Visual-Words (BoVW) \cite{peng2016bag} or the first-order statistics Vector of Locally Aggregated Descriptors (VLAD) \cite{jegou2010aggregating}. 
Videos are classified based on their BoVW representation, using non-linear 
Support Vector Machines (SVMs) with the $\chi^2$ kernel~\cite{Wan+09}.
In addition, the above different types of descriptors are combined with the Trajectory descriptor \cite{wang_action_2011} and the Histograms of Oriented Gradients (HOG)~\cite{laptev_learning_2008}, by computing distances between their corresponding BoVW histograms and adding the corresponding kernels. 

The encoded features that result from VLAD are classified employing linear SVMs.
After the feature extraction, we follow three different approaches - in multiple levels - for the fusion of the RGB information acquired by the multiple sensors: i) feature fusion, ii) encoding fusion, and iii) score fusion. We modify the general frameworks of BoVW and VLAD in order to deal with our proposed multi-view approach for visual activity recognition.

\textit{Feature Fusion:}
In this method, the visual information is fused at an early stage where only low-level $D$-dimensional feature descriptors $\mathbf{x}_{m}^i\in\mathbb{R}^D$ have been extracted, i.e., local descriptors alongside  dense trajectory $m=1,...,M_i$, from each different sensor $i=1,...,S$.
The codebook generation approach, which is based on the k-means algorithm, is modified in order to deal with the multi-view data. 
Given a set of feature descriptors $\mathbf{x}_{m}^i$, our goal is to partition the feature set into $K$ clusters $ \mathbf{D} = [ \mathbf{d}_1,\ldots,\mathbf{d}_{K} ] $, where $\mathbf{d}_k \in \mathbb{R}^D$ is the centroid of the $k$-th cluster. These $\mathbf{d}_k$ are shared between the features of all sensors. 
Using the notation of \cite{peng2016bag}, if descriptor $\mathbf{x}_m^i$ is assigned to cluster $k$, then the indicator value $r_{m,i,k} =1$ and $r_{m,i,\ell} =0$ for $\ell \neq k$. The optimal $\mathbf{d}_k$ can be found by minimizing the objective function:
\vspace{-0.2cm}
\begin{eqnarray}
\min_{{\bold{d}}_{k\,},r_{m,i,k\;}} \sum_{k=1}^{K}  \sum_{i=1}^{S} \sum_{m=1}^{M_i} r_{m,i,k} {\Vert \mathbf{x}_m^i - \mathbf{d}_k \Vert}_2^2.
\vspace{-0.2cm}
\end{eqnarray}

\noindent Then the encoding procedure is employed for both the BoVW and the VLAD method, resulting to a representation $\mathbf{s}_{n_j}^i$ for each trajectory $n_j$ of the $j$-th video captured by sensor $i$.
The global representation $\mathbf{h}$ of the multi-view video using a sum pooling scheme is given by:
\vspace{-0.2cm}
\begin{eqnarray}
\mathbf{h} = \sum_{i=1}^S \sum_{n_j=1}^{N_j} \mathbf{s}_{n_j}^i
\end{eqnarray}
Finally, for the BoVW approach, an $L 2$ normalization scheme \cite{perronnin2010improving} is applied, while for the VLAD the intra-normalization strategy proposed in \cite{arandjelovic2013all} is followed.

\textit{Encoding Fusion:}
In this approach, a different global vector $\mathbf{h}^i$ is created by encoding the dense trajectory features using a different codebook $\mathbf{D}^i$ for each sensor $i$. 
For the BoVW encoding the multi-view fusion is applied by adding the $\chi^2$ kernels: 
\vspace{-0.2cm}
\begin{equation}
K\left(\mathbf{h}_j,\mathbf{h}_q\right)=
\sum_{i=1}^S \sum_{c=1}^{N_c} \exp\left( - \frac{1}{A_c} L\left(\mathbf{h}_j^{c,i},\mathbf{h}_q^{c,i}\right)\right),
\label{equ:encod}
\end{equation}
where $\mathbf{h}_j^{c,i}$, $\mathbf{h}_q^{c,i}$ denote the BoVW representations of the $c$-th descriptor for the $j$-th and $q$-th video respectively captured by sensor $i$, and $A_c$  is the mean value of $\chi^2$ distances $L(\mathbf{h}_j^{c,i},\mathbf{h}_q^{c,i})$ between all pairs of training samples from a specific sensor $i$. 
On the other hand, for the VLAD encoding a simple concatenation of the vectors that correspond to the different sensors is applied as follows: $\mathbf{h} = [\mathbf{h}^1, \ldots , \mathbf{h}^S]$.

\textit{Score Fusion:}
For a given sensor $i$ a different SVM is trained for all employed classes and obtains the probabilities $\mathbf{P}^i$ as described in \cite{chang2011libsvm}. Then a softmax normalization is applied to each sensor's SVM probabilities. For the fusion of the different sensor output probabilities an average fusion is employed: $\mathbf{P}=\frac{1}{S}\sum_{i=1}^S \mathbf{P}^i$. Finally, the class with the highest fused score is selected, following an one-against-all approach.

\subsubsection*{Distant Speech Recognition}

In order to ensure a natural communication between humans and robots in an HRI system, it is essential to incorporate a speech recognition module. A considerable amount of distance between the robot and the users imposes the need to employ a distant speech recognition system (DSR)~\cite{wolfel2009,rodomagoulakis2017} that will have to efficiently address challenging problems, such as noise and reverberation~\cite{ishi2008robust}. 
Especially when children are the end users that play and interact with the robots, the problem of speech recognition becomes much more challenging because of the special characteristics of children voices and the difficulty to acquire quality data. 

In out setup, the microphone arrays distributed in space are employed for the DSR task. Children can use a set of utterances adopted for the specific context of the employed use-cases to communicate with the robots, thus our speech recognition system is grammar-based. A continuous system would require a large amount of children data to be collected; that was unfeasible in our case. Also, a grammar-based speech recognition system is adequate to 
fulfill the requirements of the considered use-cases. The employed language is Greek, and the set of utterances contains children possible answers in some games and some general purpose speech utterances.

The DSR system is able to detect and recognize the spoken utterances at any time, namely it is always-listening. Since speech is usually corrupted by reverberation, noise or other non-speech events, robustness is achieved via beamforming of the far-field signals and adaptation of the acoustic models.

More into the details, a sliding window of $2.5 $ sec duration 
with a $0.6$ sec shift is used in order to process speech in time frames. A custom module has been developed and integrated in Robot Operating System (ROS), which allows raw audio processing from the Kinect microphones. Each speech frame is first denoised with a simple delay-and-sum beamforming applied on each available 4-channel Kinect array:  The insertion of delays to the different microphone signals $a_n(t)$, allows us to align them appropriately, so as to enhance speech coming from a specific direction. For uniform linear arrays with $N$ microphones, which is also our case, if the desired direction is denoted by $\phi$, the time-delay to be applied to each microphone is
\begin{equation}
\tau_n = \frac{(n-1)d\cos{\phi}}{c},
\end{equation}
where $c$ is the speed of sound and $d$ the space between microphones. The beamformed signal is denoted by: 
\begin{equation}
y(t)=\frac{1}{N}\sum_{n=1}^{N}a_n(t-\tau_n)
\end{equation}

The denoised signal is then fed to the DSR module where
we enforce recognition of one of the pre-defined sentences.
Regarding acoustic modeling, Gaussian Mixture Models and Hidden Markov Models built on cross-word tri-phone models have been trained using standard Mel-Frequency Cepstrum Coefficients-plus-derivatives features on the Logotypografia database.
The Logotypografia~\cite{logotypografia} database contains clean, close-talk speech in Greek. Thus, we artificially distort the database by convolving the clean speech with room impulse responses and adding white Gaussian noise in order to match the far-field condition~\cite{rodomagoulakis2017}. Maximum likelihood linear regression 
(MLLR) adaptation is employed to transform the means of the Gaussians on the states of the models, aiming to reduce the mismatch between the initial model and the adaptation data~\cite{Young2006}.

\subsection{System Architecture}
In this subsection we describe the backbone of the perception system: the hardware architecture, the interconnection and communication between the different modules, and how the flow of the interaction is managed.

The perception modules are integrated in the full perception system based on the following hardware architecture. 
The system runs on four distributed interconnected machines, three of which run the Linux operating system and the ROS, and one the Windows Operating System. Each of the three Linux machines is connected with a Kinect V2 sensor which provides raw data (i.e., color, depth, and audio). The Windows machine is also connected to a Kinect V2 sensor, and using the Microsoft SDK Kinect V2 API provides additional skeletal and tracking information. A touch screen is also connected to the Windows machine and sends feedback to the dialog module about the choices of the children. The main data processing of the perception modules takes place on each of the three Linux machines, while the multi-view fusion is handled in one of the Linux machines.

Streaming of data and communication between the modules of the system flows via events that are transmitted through the TCP/IP broker which runs in the Windows machine and is provided by the IrisTK framework~\cite{skantze2012iristk}. Under the IrisTK paradigm, we divide events in three classes:

\begin{itemize}
	\item \textit{Sense}: events that include information about what the sensors of the system perceive
	\item \textit{Action}: events that order an actuator (i.e., a robot) to do something
	\item \textit{Monitor}: background events that contain feedback information about the actions of the system (e.g., when a robot has ended speaking) 
\end{itemize}

\textcolor{black}{Similarly, the architecture of the system was designed based on the \textit{Sense - Think - Act} principle~\cite{gat1998three}, as shown in Figure~\ref{fig:babysystem}. The multi-sensor setup of the system represents the Sense part, while the perception modules are classified into the Think principle. Finally, the multiple robots belong to the Act part of the architecture. This three-layer architecture allows for high-level modularity, in the sense that the different layers can be replaced/modified without affecting the others.}

The broker, along with the dialog management module which will be described next, acts as a central unit that receives events from all system modules and distributes them accordingly to the appropriate modules.
\textcolor{black}{In addition to the high-level modularity offered by the three-layer architecture, this unit offers an extra fine-grained modularity allowing modules to be easily removed or added in the architecture by simply defining the sets of events that the module should perceive or send back to the broker.}

The dialog manager is the central module of the system and models the flow of the interaction between the user and the system. The interaction is modeled using Harel states~\cite{harel1987statecharts}- states that can be hierarchically structured, can be executed conditionally, and contain parameters that alter the flow and the transitions between them. In addition, states can be called as functions, which means that the flow of the execution will continue to the caller state, after the callee has finished his execution.

\begin{figure}[t!]
\centering
\includegraphics[width=0.48\textwidth]{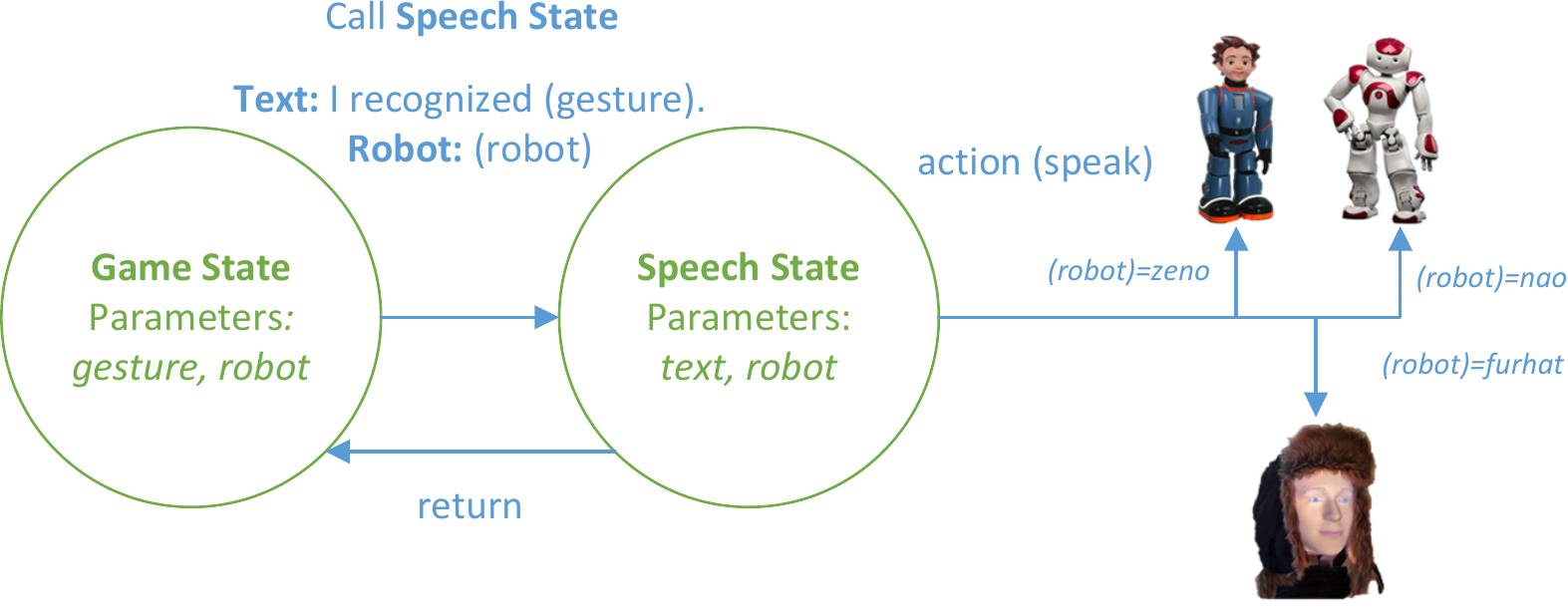}
\caption{The ``Speak Action State'' employed for announcing a gesture during the interaction.}
\vspace{-0.5cm}
\label{fig:flow_ex}
\end{figure}

For the design and development of the statechart that models the dialog, we have included what we call ``action states'', i.e., states that act as a mediator between the core dialog flow and the robots. These action states contain the information that is needed to instruct the robots of the system to perform an action, and include the robot as an additional state parameter. As a result, the core dialog flow is decoupled from robot-specific details, and we avoid defining multiple similar states for different robots. This extension also gives us the capability to easily include new robots to the dialog flow by adding the robot-specific details in the action states and handling the event on the robot side. An example can be seen in Figure \ref{fig:flow_ex} where a state in the core dialog flow  ``calls'' the ``speak action state'', including the robot that is needed to speak as an additional parameter.

From the three robots that our multi-robot system uses, the Furhat robot head and the Avatar are already integrated in the IrisTK framework. For the NAO and Zeno robots we developed intermediate APIs that we use for communication between the robots and the dialog.


\myclearpage

\section{Use-cases for CRI}
\subsection{Tasks Description}
A set of scenarios has been designed in order to highlight the capabilities of the system during an amusing and educative multimodal interaction between children and robots. As it has been explained extensively, our integrated system perceives 
various events that occur during the interaction, such as children speech and activities, along with children locations in the room, and tracking of objects. Each task focuses on different technologies and combines them appropriately to create a smooth interaction. Children are asked to complete the following tasks-games: i) ``Show me the Gesture'', ii) ``Express the Feeling'', iii) ``Pantomime'', iv)  ``Assembly Game'', v) ``Form a Farm''.

In the first task, ``Show me the Gesture", a child interacts with the robot via gestures and speech. The robot requests from the child to perform a gesture that usually denotes a meaning and tries to recognize it. It then asks the child for confirmation of the recognition. The different gestures of this game are: i) stating an agreement, ii) calling the robot to come closer, iii) asking the robot to sit down, iv) pointing an object in the room, v) asking the robot to stop, and vi) drawing a circle in the air. Except from the first gesture that is usually performed by nodding, the rest are hand gestures. The children are allowed to gesture spontaneously, as they would do when interacting with another human.

The ``Express the Feeling" game motivates children to reveal their feelings using both their face and their body during an entertaining interaction with the robot. In this game, the child selects one of the cards that are depicted on the touch screen and expresses the chosen feeling. The emotions included in this game are happiness, sadness, fear, anger, surprise, and disgust. After the child's reaction, the robot also expresses the same feeling using its body and face. 

``Pantomime" is a popular game, during which, one person mimes a handwork and the other figures out the depicting handwork. The child can use the whole body to mimic an activity and interact extensively with the robot. The robot and the child repeatedly swap the roles of the mime and the guesser. The twelve activities used in this game are the following: i) cleaning a window, ii) driving a bus, iii) hammering a nail, iv) swimming, v) working out, vi) dancing, vii) reading a book, viii) digging a hole, ix) playing the guitar, x) wiping the floor, xi) dancing, and xii) ironing a shirt. 

For the ``Assembly Game" one or more children are asked to complete an assembly under robot supervision. Six interconnectable 3D printed bricks of different lengths are used to create rectangles and squares. The bricks are placed on a table in front of the child, with the robot standing close by. The child is responsible for the manipulation of the assembly subcomponents, while the robot provides instructions and feedback. If the child correctly completes a connection, the robot congratulates the child and proceeds to give the next instruction. If the child makes a mistake, however, the robot will attempt to recognize this mistake and will react accordingly. Aside from verbal instructions, the robot also looks and points at the bricks that it refers, for clarity.

The ``Form a Farm" game is a multi-party game scenario involving two roles that can be interchanged and be equally played by both the robot and the children, aiming to entertain, educate, but also establish a natural interaction between all parties. The game involves two different roles. These roles, the picker and the guesser, can be equally played and interchanged between the children and the robot. The picker chooses an animal and utters characteristics of it. The guesser has to guess the picked animal.
The interaction proceeds as follows:
At first, the robot chooses a random animal and the human players take turns guessing the chosen animal. In case of a wrong guess, the robot reveals more characteristics of the animal (animal color, number of legs, animal class, e.g., mammals, reptiles). In case of correct identification of the animals, the robot asks the children to properly place the animal in a farm with some distinct segmented areas, which appears in a touch screen in front of them. 
In the second round, the roles are reversed: children discuss and pick an animal, and reveal one characteristic. The robot then tries to guess the picked animal. If the robot guesses correctly, the children are again asked to place the animal in the farm, otherwise they reveal more characteristics of the animal, one at at time. The game continues by interchanging the role of the guesser between children and robots in each round. The game features a total of 19 animals, and their characteristics belong to five different classes: color, size, species, number of legs, and a distinctive property, i.e. for the dog: ``it's the human's best friend''.

\begin{table*}[!th]
\small
\centering
\begin{tabular}{|l|c|c|c|c|c|c|c||c|c|c|}  \cline{2-11}
\multicolumn{1}{c|}{}  & Distant Speech & Detect  & Speaker & \multicolumn{2}{c|}{Visual Act. Rec.} & Touch & Behavioral & \multicolumn{3}{c|}{Robots}\\ \cline{5-6} \cline{9-11}
 \multicolumn{1}{c|}{}  & Recognition & \& Track & Loc. & Gesture & Action & screen  & Generation & NAO & Furhat & Zeno  \\ \hline
Show me the Gesture & \checkmark & & \checkmark & \checkmark & & & \checkmark  & \checkmark  & \checkmark  & \checkmark \\ \hline
Pantomime & \checkmark & &  \checkmark  & & \checkmark  & \checkmark & \checkmark & \checkmark & &  \\ \hline
Assembly Game & & \checkmark & \checkmark & &  & & \checkmark & \checkmark &  & \checkmark \\ \hline
Form a Farm & \checkmark & & \checkmark &  \checkmark &  & \checkmark & \checkmark & \checkmark & \checkmark & \checkmark \\ \hline
Express the Feeling &  & &  &   &  & \checkmark & \checkmark &  & \checkmark & \checkmark \\ \hline
\end{tabular}
\caption{\textcolor{black}{Used ChildBot technologies in each use-case scenario and the eligibility of each robotic agent for participating.}}
\label{tab:modules}
\end{table*}

The aforementioned tasks aim to create a proper framework for multimodal communication between children and robots, as it happens between humans. \textcolor{black}{This way, the tasks demonstrate the capabilities of the system, give some examples of how ChildBot can be used, and can be employed for system evaluation. Taking into consideration also the fact that the tasks are destined for children, the use-cases were designed under the supervision of psychologists and pilot studies with eight children were conducted, leading to the presented scenarios.} Even though each task focuses on one of the system perception technologies, more than one modules are used in parallel. \textcolor{black}{ In Table~\ref{tab:modules}, the used modules are summarized along with the eligibility of the robots to participate in each task.}

\label{section:usecases}
\subsection{Database}
\label{section:data}

\begin{figure}[!t]
\centering
\resizebox{.9\linewidth}{!}{
\includegraphics[]{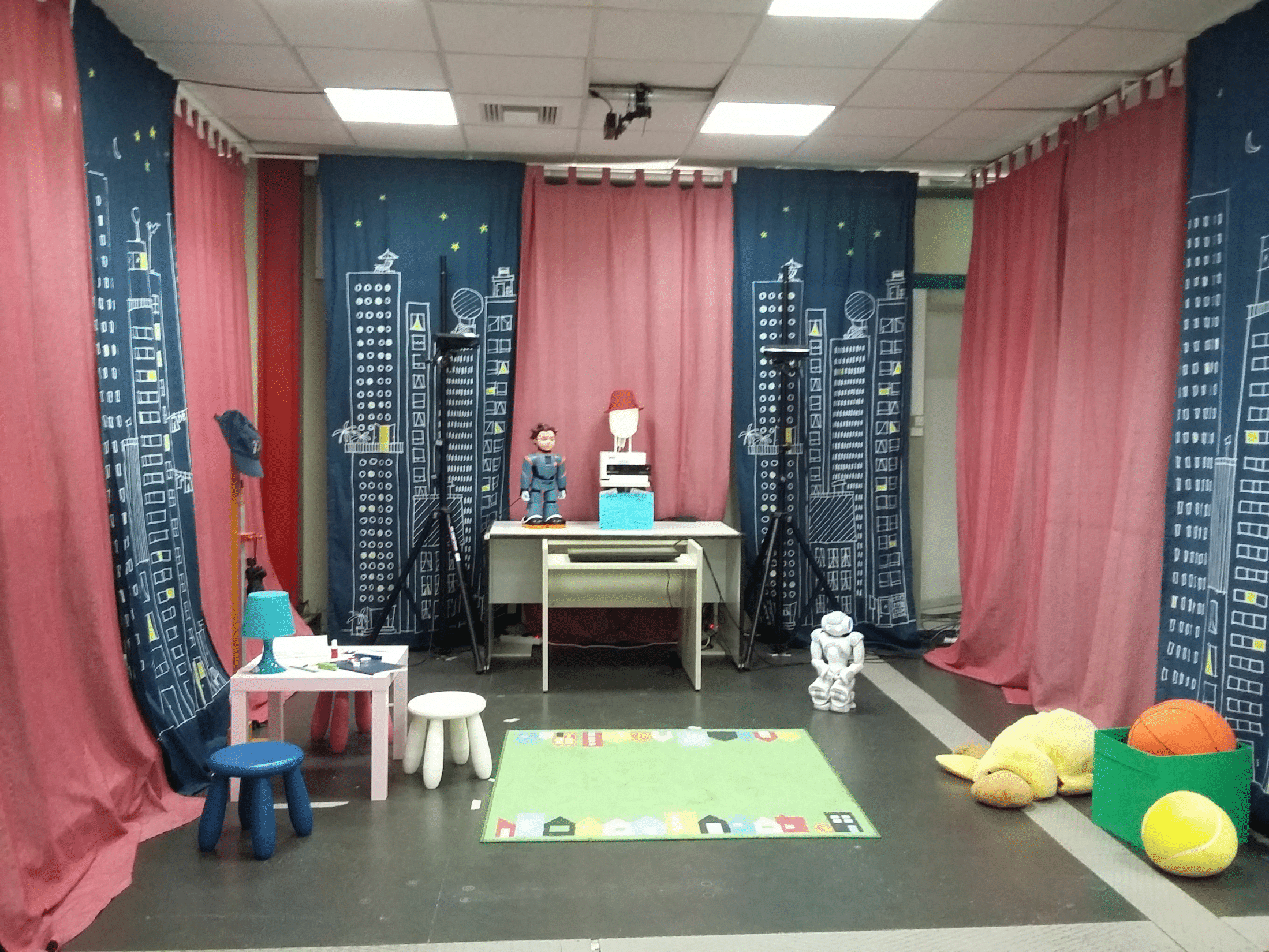}
}
\caption{Data collection room and experimental setup.}
\label{fig:collection_room}
\end{figure}
 
Real data obtained through HRI prove to be especially important during the process of developing a system, from the training to the evaluation stage. Such data contribute to an adaptation of the system to actual circumstances and human spontaneous behavior. Thus, an extensive data collection has taken place with the participation of a pool of 52 children, aged from six to eleven years old, in a specially designed room and in a school classroom.

Most of the data have been collected in a room that resembles a child's room and where the robotic agents and the sensors have been located, as it is presented in Figure~\ref{fig:collection_room}. There, the data collection has been carried out in two phases. In the first one, children data have been recorded while performing certain actions and uttering certain phrases that are expected to arise throughout the interplay between them and the robots, in a strictly controlled way, when asked to do so. These data will be referred to below as \textit{development data} since they have been used for the development of the system. In the second phase, the data have been collected during the experimental procedure where children interact with robots without any interruption or other people's intervention. The latter data will be referred to as \textit{use-case related data}. Both types of data are equally important for CRI, as the first one is indispensable for training the perception modules on data that are relevant to use-cases, while the second one is essential for the testing of the behavioral monitoring software during CRI. Table~\ref{tab:Data} presents the most important recorded events during the two phases and the total number of their occurrences.

The information we collected during the data collection includes Full HD ($1920\times1080$) RGB and depth ($512\times424$) video streams from all four Kinect cameras, running at 30fps, as well as raw audio from the microphone array embedded in each Kinect sensor. By exploiting the Kinect v2 API we have also captured the following streams from the frontal Kinect sensor: (a) Skeletal information both in 2D (image) and 3D (world) coordinates; (b) Bounding boxes from face detection, facial landmarks and a facial 3D mesh. 
 
For the development data, 28 children have participated by performing seven gestures and twelve pantomimes, and uttering 40 phrases from a vocabulary of 120 phrases. This phase is crucial for developing the perception models and adapting them to children, since they focus on speech, gestures, and actions relevant to the use-cases. Specifically, children are more spontaneous and expressive than adults and their speech is usually brief and low-voiced. Thus, in order to test the performance of ChildBot modules, it is necessary to have a plethora of children activities and utterances.
Moreover, adults' data have been collected to augment the data related to the use-cases as well as to validate and highlight the need of children data \textcolor{black}{for enhancing performance in the perception models.}

  \textcolor{black}{As far as the use-case related data are concerned, 31 children with an average of 8.6 years old, 10 girls and 21 boys, had been chosen randomly from a set of volunteers that met our team in a dissemination event. All children, from six to eleven years old, spoke Greek and were able to read and write. Each child accompanied by his/her parents entered the specifically designed room and was introduced to the robots by a researcher. The child got familiarized with the room and the robots while the researcher explained the structure of the procedure and the tasks as they were presented above. 
Afterwards, the parents and the researcher exited the interaction space and the child played the individual games with the robots. After completing the individual interaction, a second child (which had completed the same interaction previously) entered the space and collaborated with the other child while playing the ``Form a Farm" task. In cases where there was no second child available, an adult took its place. However, these data were removed from the subsequent evaluation.
Finally, after completing the procedure, the children were asked to complete a questionnaire that included subjective statements regarding their experience. The questionnaire will be described and discussed in Section~\ref{subsection:subjeval}.}

\textcolor{black}{The above procedure has been approved by the Ethics Committee of the Athena Research Center, and also includes a consent form that had been sent by email to the parents before the experiments. In addition, all experiments have been supervised by an experienced child psychologist.
}

\begin{table}[!tb]
\centering
\begin{tabular}{|c|l|c|}  \hline
  Collected Data & Event Type & \# of Events\\ \hline \hline
 & Utterances & $977$ \\ \cline{2-3}
Development & Gestures & $196$ \\\cline{2-3}
Data  & Pantomimes & $336$ \\ \hline \hline
Use-case & Utterances & $641$ \\\cline{2-3}
Related & Gestures & $143$ \\\cline{2-3}
Data  & Pantomimes & $109$ \\ \hline
\end{tabular}
\caption{Statistics of the most important child activities during the data collection. }
\label{tab:Data}
\end{table}

\begin{figure*}[!t]
\centering
\subfigure[``Show me the Gesture'', Kinect\#$1$]{\includegraphics[height=0.19\textwidth]{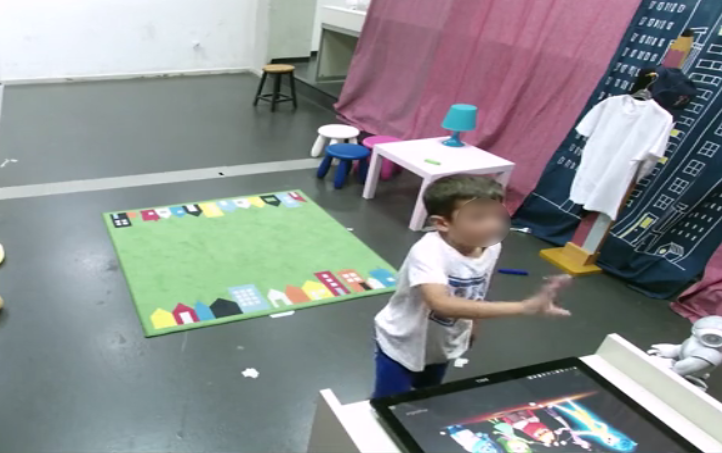}}
\subfigure[``Pantomime'', Kinect\#$2$]{\includegraphics[height=0.19\textwidth]{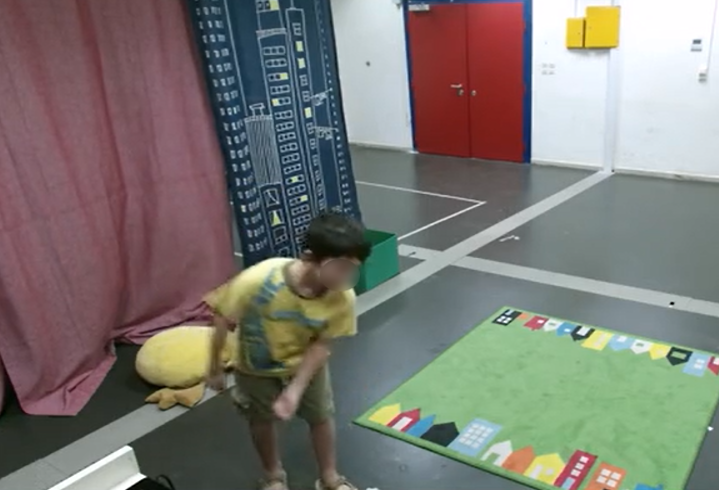}}
\subfigure[``Form a Farm'', Kinect\#$3$]{\includegraphics[height=0.19\textwidth]{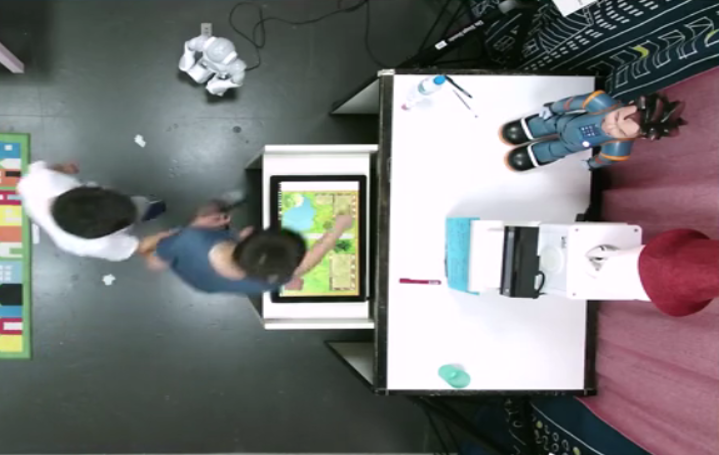}}
\subfigure[``Express the Feeling'', Kinect\#$4$]{\includegraphics[height=0.19\textwidth]{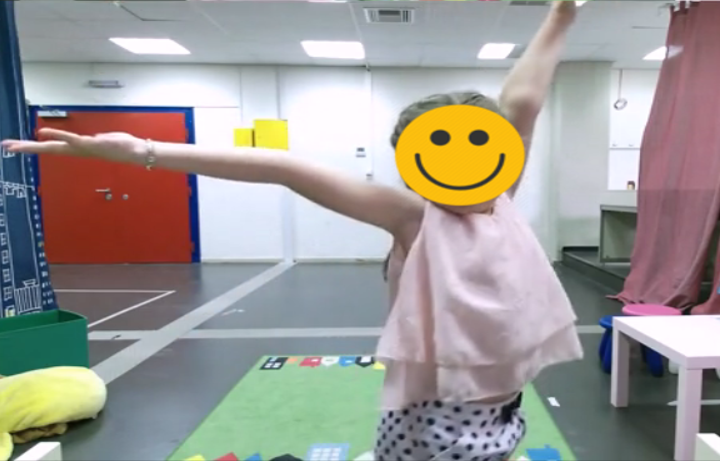}}
\caption{The four different use-cases that took place in our laboratory, each one presented from one the four different camera viewpoints of ChildBot.}
\label{fig:sde}
\end{figure*} 

\begin{figure}[b]
	\centering
	\includegraphics[width=0.5\linewidth]{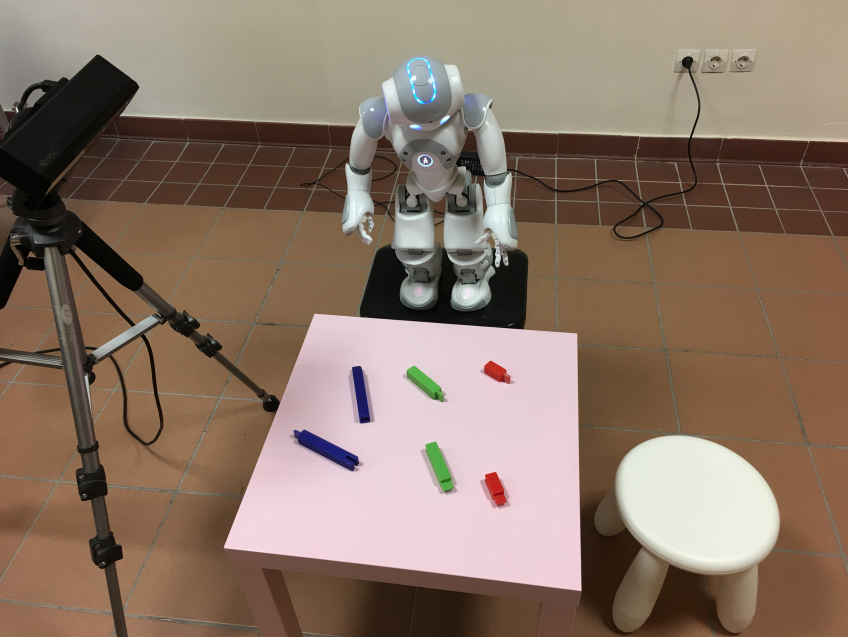}
	\caption{Setup of the ``Assembly Game'' at a Greek primary school.}
	\label{fig:Assembly}
\end{figure}

The use-case related data regarding the ``Assembly Game'' were collected in a Greek primary school where 21 students, 9-10 years old, participated either individually or in groups of five. For this task, a single Kinect camera and one robotic agent (NAO robot) have been chosen as a lightweight version of the system to accommodate the educational process. Such a version can be easily installed in a typical classroom and help the teacher to give a vivid lesson through a CRI experience (Figure~\ref{fig:Assembly}).

\myclearpage

\section{CRI Evaluation}
\textcolor{black}{
Each perception module of the ChildBot system has been evaluated by measuring its performance in efficient multi-modal scene understanding, using the collected data. In addition, we have performed a preliminary user experience study in order to assess how the children interact and perceive the system, and collect insights towards carrying out a more complete subjective evaluation in the future.}

\subsection{Perception module evaluation}
\subsubsection*{Audio-Visual Active Speaker Localization}

 The evaluation results of audio-visual speaker localization are presented in Table~\ref{tab:sloc}.
For audio-only speaker localization, the employed metrics are Pcor (Percentage correct) which is the percentage of correct estimations (deviation from ground truth less than $0.5$m) over all estimations, RMSE (Root Mean Square Error) between the estimation and the ground truth, and RMSEf (RMSE for estimations
with error less than $0.5$m - i.e., 'fine errors') .
For audio-visual speaker localization, since person locations are estimated by the Kinect skeleton, the problem is essentially transformed into an active speaker localization problem. Thus, evaluation is performed in terms of correct speaker estimation, where Pcor is used, denoting the correct speaker estimations over all estimations.
Audio-only localization does not perform sufficiently well yielding a Pcor of $45\%$, but the average RMSE is $60$cm, meaning that the average source localization error is $60$cm which is not very large.
If both audio and visual information are used, then the active speaker localization performance is boosted to a Pcor of $86\%$.

\begin{table}[!h]
\centering
\begin{tabular}{|l|c|c|c|} \hline
 \multicolumn{3}{|c|}{Audio source }  & \multicolumn{1}{c|}{Audio-visual active} \\ 
  \multicolumn{3}{|c|}{localization}  & \multicolumn{1}{c|}{speaker localization} \\ \hline
\multicolumn{1}{|c|}{Pcor} &  \multicolumn{1}{c|}{RMSE}   &  \multicolumn{1}{c|}{RMSEf}   &  \multicolumn{1}{c|}{Pcor}  \\ \cline{1-4}
$45.51\%$ & $0.60$m & $0.35$m & $85.58\%$ \\ \hline
\end{tabular}
\caption{Evaluation of the audio-visual active speaker module.} 
\label{tab:sloc}
\end{table}
\subsubsection*{6-DoF Object Tracking}

For object tracking, we have performed both an objective evaluation and a subjective evaluation, in order to assess the performance of the 3D visual tracking module.

During the objective evaluation, because it is difficult to annotate and obtain ground truth poses for 3D tracking, we have placed two static objects on a table, along with obstacles in order to add occlusions, and have moved a camera around the objects with sudden movement bursts, in order to establish the robustness of the tracker. We have compared our method with an SDF (Signed Distance Function) tracker \cite{Ren+14}. Our results have showed that, although the SDF tracker has produced a steadier output than our tracker in cases of partial occlusions and slow camera movements, when we have introduced sudden jolts and full occlusion, the SDF tracker has been unable to continue tracking and  has failed, without recovering, even when the normal conditions (no occlusion-jolts) were restored. On the other hand, our tracker has been able to successfully track the object with low error, even under full occlusion and fast movements, and recover in rare cases where the tracking has been lost, without the need for reinitialization, proving its robustness. More information on these objective evaluations can be found in \cite{Hadfield2018}.

During the subjective evaluation, we used the NAO robot as a supervisor for the Assembly Game which is described in Section 4. 
From the 21 participants, 6 played the game on their own, while the remaining children played in groups of 5. 
The children were required to complete two different rectangles and one square, by choosing and connecting items from 6 different brick objects. 

In Table \ref{tab:assembly_results} we can see the results from the experiments in the form of statistics for the different assemblies. \textcolor{black}{We present the percentage of total and required connections that the system recognized within a time interval of 5 sec and 20 sec, which is referred to as identification time.
The term ``total connections" includes both correct connections that the child completed and mistaken connections, while the term ``required connections" refers only to the correct connections needed to complete the assembly.}

\begin{table}[h]
\centering
\resizebox{0.98\linewidth}{!}{
\begin{tabular}{l|c|c|c|c|}
\cline{2-5}
              & \multicolumn{2}{c|}{Total Connections}  & \multicolumn{2}{c|}{Required Connections} \\ 
              & \multicolumn{2}{c|}{(Recall $\%$)}  & \multicolumn{2}{c|}{(Recall $\%$)} \\ \hline
\multicolumn{1}{|c|}{Identification Time}  &  5s & 20s & 5s & 20s \\ \hline
\multicolumn{1}{|c|}{Rectangle} &  70.00 & 80.00 & 50.0 & 56.25 \\
\multicolumn{1}{|c|}{Square} & 39.39 & 57.58 & 43.24 & 59.46 \\
\hline 
\end{tabular}
}
\caption{Statistics about the performance of the 6-DoF object tracking employed in ``the Assembly Game''.}
\label{tab:assembly_results}
\end{table}

\subsubsection*{Visual Activity Recognition}

\begin{table}[!tb]
\renewcommand{\arraystretch}{1}
\centering
\begin{tabular}{|c|l|c|c|c|} \cline{3-5}
 \multicolumn{1}{c}{} &   \multicolumn{1}{c|}{}  & \multicolumn{3}{c|}{Training group} \\ \cline{2-5}
\multicolumn{1}{c|}{}  & Testing   & & & \\ 
\multicolumn{1}{c|}{}  &  group   & Adults   &  Children  & Mixed  \\ \hline
 Gesture  & Adults
& $92.19$ & $62.08$ & $\mathbf{95.10}$ \\  \cline{2-5}
Recognition & Children 
 & $56.25$ & $\mathbf{83.80}$ & $80.09$  \\ \hline
Action & Adults 
 & $\textbf{87.36}$ & $72.53$ & $86.26$ \\  \cline{2-5}
Recognition & Children 
& $56.51$ & $\textbf{74.46}$  & $74.26$  \\ \hline
\end{tabular}
\caption{Evaluation of the activity recognition modules after the fusion of different camera scores using MBH features and BoVW encoding.}
\label{tab:pres2}
\end{table}

\begin{table*}[!tb]
\centering
\scriptsize
\scalebox{1.1}{\begin{tabular}{|l|c|c|c|c||c|c|c|c|c|c|} \cline{2-11}
\multicolumn{1}{c|}{} & \multicolumn{10}{c|}{Development Data} \\ \cline{2-11}
           \multicolumn{1}{c|}{}  & \multicolumn{4}{c||}{Single Camera} & \multicolumn{6}{c|}{Fusion}\\\hline
Features       &  Kinect \#$1$   & Kinect \#$2$   & Kinect \#$3$  & Kinect \#$4$  & \multicolumn{2}{c|}{Features} & \multicolumn{2}{c|}{Encodings} & \multicolumn{2}{c|}{Scores} \\ \cline{2-11}
 &  \multicolumn{4}{c||}{BoVW} & BoVW & VLAD & BoVW & VLAD & BoVW & VLAD \\ \hline
HOF  & 70.83 & 70.37 & 69.21 & 63.43 & 71.76 & 74.07 & 77.78 & 81.48 & 75.93 & 81.94 \\
MBH  & 76.85 & 67.82 & 68.29 & 65.28 & 76.39 & 76.85 & 81.02 & 81.48 & 82.87 & 83.80 \\ 
Traj.+HOG+HOF+MBH & $\mathbf{77.78}$ & 73.84 & 73.61 & 75.00  & 81.48 & 82.87 & 82.87 & 83.80 & 82.87 & $\mathbf{85.19}$\\\hline 
\multicolumn{1}{c|}{} & \multicolumn{10}{c|}{Use-case Related Data} \\ \cline{2-11}
\multicolumn{1}{c|}{}  & \multicolumn{4}{c||}{Single Camera} & \multicolumn{6}{c|}{Fusion}\\\hline
Features       &  Kinect \#$1$   & Kinect \#$2$   & Kinect \#$3$  & Kinect \#$4$  & \multicolumn{2}{c|}{Features} & \multicolumn{2}{c|}{Encodings} & \multicolumn{2}{c|}{Scores} \\ \cline{2-11}
 & \multicolumn{4}{c||}{BoVW} & BoVW & VLAD & BoVW & VLAD & BoVW & VLAD \\ \hline
HOF  & 56.92 & 54.49 & 57.10 & 51.97  & 54.56 & 71.61 & 58.01 & 74.73 & 63.26 & 74.83\\
MBH  & 62.70 & 56.47 & 60.15 & 54.25 & 65.32 & 72.70 & 67.72 & 72.52 & 66.73 & 72.72\\
Traj.+HOG+HOF+MBH  & 57.96 & 54.08 & $\mathbf{67.03}$ & 59.16 & 61.51 & 69.85 & 63.38 & $\mathbf{73.95}$ & 64.82 & $\mathbf{73.35}$ \\ \hline
\end{tabular}}
\caption{Average classification accuracy (\%) for the employed 8 gestures. Results on both development and use-case related data are shown for the different features, encoding and fusion methods of the activity recognition module.}
\label{tab:gres}
\end{table*}

\begin{table*}[!tb]
\centering
\scriptsize
\scalebox{1.1}{\begin{tabular}{|l|c|c|c|c||c|c|c|c|c|c|} \cline{2-11}
\multicolumn{1}{c|}{} & \multicolumn{10}{c|}{Development Data} \\ \cline{2-11}
           \multicolumn{1}{c|}{}  & \multicolumn{4}{c||}{Single Camera} & \multicolumn{6}{c|}{Fusion}\\\hline
Features      &  Kinect \#$1$   & Kinect \#$2$   & Kinect \#$3$  & Kinect \#$4$  & \multicolumn{2}{c|}{Features} & \multicolumn{2}{c|}{Encodings} & \multicolumn{2}{c|}{Scores} \\ \cline{2-11}
 & \multicolumn{4}{c||}{BoVW} & BoVW & VLAD & BoVW & VLAD & BoVW & VLAD \\ \hline
HOF  &  68.31 &  56.31 &  48.62 &  53.85 & 66.77 & 67.08 & 68.00 & 69.23 & 68.62 & 75.50\\ 
MBH  &   70.77 & 60.92 & 61.85 & 55.22 & 76.00 & 76.69 & 76.92 & 76.92 & 74.46 & 76.50\\ 
Traj.+HOG+HOF+MBH & $\mathbf{73.85}$ &  63.38  & 60.00  & 61.45 & 75.08 & 76.92 & 77.23 & 77.85 & 75.08 & $\mathbf{79.00}$\\\hline 
\multicolumn{1}{c|}{} & \multicolumn{10}{c|}{Use-case Related Data} \\ \cline{2-11}
\multicolumn{1}{c|}{}  & \multicolumn{4}{c||}{Single Camera} & \multicolumn{6}{c|}{Fusion}\\\hline
Features      &  Kinect \#$1$   & Kinect \#$2$   & Kinect \#$3$  & Kinect \#$4$    & \multicolumn{2}{c|}{Features} & \multicolumn{2}{c|}{Encodings} & \multicolumn{2}{c|}{Scores} \\ \cline{2-11}
 & \multicolumn{4}{c||}{BoVW} & BoVW & VLAD & BoVW & VLAD & BoVW & VLAD \\ \hline
HOF  & 46.34 & 46.19 & 25.50 & 47.70 & 63.08 & 61.02 & 49.87 & 56.17 & 52.59 & 57.99\\
MBH  & $\mathbf{61.42}$ & 46.28 & 31.59 & 45.57 &  $\mathbf{70.25}$ & 67.97 & 57.70 & 59.04 & 62.18 & 62.49\\
Traj.+HOG+HOF+MBH   & 52.59 & 46.74  & 36.62  & 48.16 & 63.52 & $\mathbf{69.37}$ & 60.75 & 61.55 & 55.00 & 64.90\\ \hline
\end{tabular}}
\caption{Average classification accuracy (\%) for the employed 13 pantomimes. Results on both development and use-case related data for the different features, encoding and fusion methods of the activity recognition module are depicted.}
\label{tab:gres2}
\end{table*}

Firstly, we have examined if the age group of the participants (adults or children) has an impact on the accuracy of visual activity recognition. For both visual activity recognition tasks, we trained separate models using as training set: a) children, b) adults, c) mixed (both adults and children) and as testing set: a) children, b) adults. In Table~\ref{tab:pres2}, it can be noticed that the use of children training data is imperative for achieving high accuracy in children activity recognition, irrespectively of the task. This result justifies our choice for collecting development data from children movements. On the other hand, recognition models trained on mixed age groups perform better for adult gesture recognition since the diversity with which children perform the gestures accommodates the generalization of the model. For the action recognition task, children employ a wider range of different movements that adults do not use as they act stereotypically, and the mixed age training models perform worse than the adults models.

Furthermore, Tables~\ref{tab:gres} and~\ref{tab:gres2} summarize the evaluation of gesture and action recognition respectively, for several combinations of different features, encodings in both the single-view case and multi-view case along with the multi-level fusion. Also, the recognition models have been trained on children development data and tested on both development and use-case related data separately, using the leave-one-out cross-validation approach.

Specifically, Table~\ref{tab:gres} presents average accuracy results (\%) for the 7 gestures and a background model. Results indicate that the best multi-view model outperforms the best single-view model by about 7\%, which underlines the need of a multi-view system for unrestrained CRI. For the development data, it can be seen that the combination of different types of features performs better than HOF and MBH features individually. Among the single-sensor cases, Kinect\#$1$ (right side view) performs best as most of the kids are right-handed and they stood at approximately the same location while performing the gestures. The best recognition accuracy of $85.19\%$ is noticed for the fusion in the final step of the procedure with the VLAD encodings and the feature combination. As far as the use-case related data are concerned, the accuracy for the single streams is moderately lower than the previous ones, which reveals the difficulty that children faced while they were trying to perform the gesture spontaneously. Also, as the children stand at a completely different location, usually closer to the cameras, the best single stream result appears for Kinect\#$3$ (floor plan view). Regarding the fusion of the different streams, recognition performance is slightly better for the encodings fusion than the scores fusion, and it approaches $74\%$.
More generally, VLAD encodes more effectively the visual information than the BoVW, since it contains rich information about the distribution of the visual words. Finally, we have to note that, as nodding requires a gentle movement, it is usually confused with the background movement.

In order to verify the appropriateness of the proposed visual activity recognition system in more challenging tasks, we evaluate the visual activity recognition system for the pantomimes.
Table~\ref{tab:gres2} presents the average accuracy results (\%) for the 12 pantomimes as well as the background model. The fusion of the single-view information enhances remarkably the performance of the recognition, as was observed in the gesture case too. The highest accuracy for testing on development data appears with VLAD encodings in scores fusion, since the visual information in these data is more consistent compared to use-case related data, e.g. similar time duration of a pantomime or similar children locations in the room. Regarding the single-view case, in both types of data, the right side view Kinect\#$1$ appears to be the best perspective for the trained models. It can be noticed that in use-case related data MBH yields slightly better results than feature combination.
Moreover, feature fusion, i.e. fusion  of the information at an early step of the entire procedure results in the best performance regardless of the type of the encodings.    

In conclusion, the accuracy of the visual activity recognition is lower in use-case related data since children act more spontaneously while they move around the room and interact freely with the robots. Furthermore, due to the fact that the variation of the visual information in the pantomime task is larger than in the gesture task, the early fusion of the features performs better for the pantomime while the scores fusion is satisfactory for the ``Show me the Gesture'' since only one camera is adequate to recognize the gesture.

\subsubsection*{Distant Speech Recognition}

Two sets of data have been employed for the offline evaluation of the DSR task: the development set and the use-case related set, both consisting of children data. As stated before, the DSR system is grammar-based, namely depending on the context of the application, and there is a specific set of commands that the users adopt in order to communicate with the robot.
Thus, we have designed one set for the ``Show me the Gesture'', the ``Express the Feeling'', and  the ``Pantomime'' games and another one for the cooperative game, i.e. the ``Form a Farm'' game. The grammar size and other statistics concerning the two datasets can be found in Table~\ref{tab:datastats}.

The development data have been used for adapting speech models and testing them. 
Results are presented in Table ~\ref{tab:sres2} in terms of word and sentence accuracies, denoted by WCOR and SCOR respectively.
Four different adaptation schemes have been tested for comparison: In the ``No-adapt" case, the employed models have been trained on the Logotypografia database which contains adult data. The available children data included in the development set have been used for testing.
In the ``Adults" case, speech models have been adapted to a small amount of adult data and tested both with adult and children data.  In the ``Children" case, data from $20$ out of $28$ participants of the development set have been used to adapt speech models globally, i.e. data from the Kinect arrays have been used to adapt a single model. The remaining $8$ participants form the testing set. The adaptation and testing has been 4-fold cross-validated. Finally, in the ``Mixed" case we have used both adult and children data to adapt the models and then test them separately on adult and children data.
\begin{table*}[t!]
\centering
\begin{tabular}{|l|c|c|c|c|c|c|} \cline{2-7}  
\multicolumn{1}{c|}{} &  \multicolumn{3}{c|}{Development set}   &  \multicolumn{3}{c|}{Use-case Related set} \\ \cline{2-7}
\multicolumn{1}{c|}{} & \multicolumn{1}{c|}{\# speakers} & \# utterances & grammar size & \# speakers & \# utterances & grammar size \\ \hline 
Single Game &  $28$   & $642$ & $75$  & $31$  & $426$ & $157$  \\ \hline 
Cooperative Game  &  $28$ & $335$ & $58$ & $9$ pairs & $215$ & $113$      \\ \hline 
\end{tabular}
\caption{Data statistics for the children DSR task.}
\label{tab:datastats}
\end{table*}

\begin{table*}[t!]
\centering
\begin{tabular}{|l|c|c|c|c|c|c|c|c|} \cline{2-9}
\multicolumn{1}{c|}{} & \multicolumn{8}{c|}{DSR-Adaptation scheme} \\ \cline{2-9}
 \multicolumn{1}{c|}{} &  \multicolumn{2}{c|}{No-adapt}   &  \multicolumn{2}{c|}{Adults}   &  \multicolumn{2}{c|}{Children}  & \multicolumn{2}{c|}{Mixed}  \\ \cline{1-9}
\multicolumn{1}{|c|}{Test}  & \multicolumn{1}{c|}{WCOR} & SCOR & WCOR & SCOR & WCOR & SCOR & WCOR & SCOR  \\ \hline \hline
Adults & $97.54$ & $91.25$ & $99.58$ & $\mathbf{98.87}$ & $96.73$ & $93.20$ & $99.50$ & $98.43$    \\ 
Children  & $79.06$ & $69.95$ & $75.31$ & $71.20$ & $97.81$ & $\mathbf{95.50}$ & $90.71$ & $82.06$   \\ \hline
\end{tabular}
\caption{Evaluation of the DSR  recognition on the development data.}
\label{tab:sres2}
\end{table*}

\begin{table*}[!ht]
\centering
\begin{tabular}{|l|c|c|c|c|c|c|} \cline{2-7}  
\multicolumn{1}{c|}{} &  \multicolumn{3}{c|}{No-adapt}   &  \multicolumn{3}{c|}{Adapt-all} \\ \cline{2-7}
\multicolumn{1}{c|}{} & \multicolumn{1}{c|}{WCOR} & SCOR & LabelCOR & WCOR & SCOR & LabelCOR \\ \hline 
Single Game & $56.68$ & $29.52$ & $55.12$  & $59.64$ & $43.77$ & $ 55.12 $            \\ \hline 
Cooperative Game  & $72.95$ & $61.02$ & $63.16$ & $78.00$ & $67.69$ & $70.51$              \\ \hline 
\end{tabular}
\caption{Average word (WCOR), sentence accuracy (SCOR), and Label accuracy (LabelCOR) (\%) for the children DSR task.}
\label{tab:sres}
\end{table*}
Speech recognition achieves satisfactory performance for adults even without adaptation. However, adaptation indeed improves performance for all cases, even when it is performed in a different age group than testing. Results indicate that the best performance is obtained for adapting and testing on the same group, which was expected. The best achieved results are $98.87\%$ for adults and $95.50\%$ for children in terms of SCOR.
The results concerning children underline the need and importance of collecting children data.  \textcolor{black}{Performance is boosted, from $75.3\%$ to $97.8\%$  for WCOR and from $71.2\%$ to $95.5\%$ for SCOR, when children data have been used for adaptation and testing, who are indeed the target group of the system}. 

All the above results were referring to the development set where data collection was controlled and guided. The use-case related set contains data in the wild, namely, while children were actually playing with the robots, their speech data were collected. Although the children had received some instructions concerning the utterances they could use, it is obvious that they were not followed in most cases. Thus, after the data annotation, new grammar sets have been formed in order to incorporate new phrases.

The use-case related data results are depicted in Table~\ref{tab:sres}, in terms of WCOR, SCOR and LabelCOR. LabelCOR refers to the percentage of correct recognition of the semantic content. For example, there can be various ways to express a negation: ``no", ``no, I don't know", ``no, I did not find it", etc. All similar phrases in terms of content are given a specific label, and after speech recognition data post-processing calculates the score of the correct recognition of labels.
Adaptation has been performed using the development set. We notice that adaptation boosts the performance, achieving a percentage of $59.64\%$ for the gesture and pantomime games and $78\%$ for the farm game in terms of WCOR. The lower percentage of gesture and pantomime games can be attributed to the larger grammar size, the distance between the speakers and the microphones, and the relatively large variations of speaker orientation.

\subsection{\textcolor{black}{User Experience Study}}
\label{subsection:subjeval}


\begin{figure*}[!t]
\centering
\includegraphics[width=0.92\textwidth,]{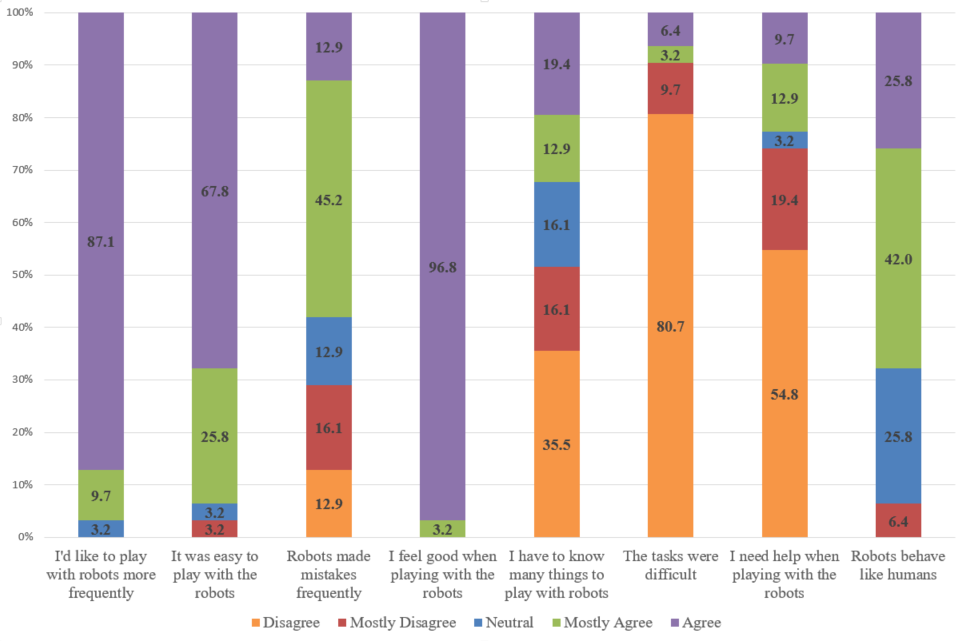}
\centering
\caption{User experience of the entire ChildBot system. After each completed interaction, children were asked to fill a questionnaire with the shown questions on a Likert-type scale from 1 to 5 labeled as shown.}
\label{fig:subj}
\end{figure*}

\textcolor{black}{We have also performed a user experience study which included a pool of 52 children, from six to eleven years old, participating in the designed interaction described in Section \ref{section:data}. The purpose of this study is to collect objective statistics and insights, as well as get a measure of the system's ability to accommodate a complete CRI.}

\textbf{Objective Statistics} Regarding the individual tasks (``Show me the Gesture'', ``Express the feeling'', ``Pantomime''), where each of the 31 kids participated alone, all of them were able to complete successfully the games. The average duration of completion, including the introduction by the robots, is 9 minutes, with a variance of 2 minutes. 

In 32\% of the cases, human (verbal) intervention was required up to two times during the experimental flow, when the children became confused or had questions about the procedure. For example, some children asked for a confirmation about what to do or needed a prompt in order to act. Such possible deviations from the designed scenario have been overcome by enabling the dialog manager to recognize these cases (e.g., if the child is silent for a moderate period of time), and getting the robots prompt the children or ask them to repeat their utterance/activity. In cases where children were expected to say something or their speech was not recognized, robots requested for repetitions up to two successive times, while in case of a child's action, the robots asked for repetition only once.

For the collaborative ``Form a Farm'' game, played by two children, it was observed that younger children faced difficulties with the rules of the game, even though primary school children are familiar with the animals of a farm. As a result, kids with ages six and seven played the game following the guidelines offered by one adult. The rest of the children played the game without any guidance. \textcolor{black}{The average duration of the game was 8 minutes}. In total, children assumed the role of the guesser for 24 rounds and found the correct answer using 2.4 guesses on average and 4 guesses maximum. On the other hand, the robot assumed the guesser role for 22 rounds and found the correct answer in 2.2 guesses on average, with a maximum of 6. Children did not manage to identify the picked animal in 4\% of the guesses, while the robot in 32\%. Generally, the children managed to guess the picked animal more easily, since the robot was programmed to always reveal more general animal characteristics in the beginning, and proceeding with more specific details.

\textcolor{black}{\textbf{User experience assessment} Regarding subjective evaluation of the experience}, children were asked to complete a questionnaire containing the subjective statements that can be seen in Figure \ref{fig:subj}. Each statement was accompanied by a 5-point Likert type ordinal scale labeled from ``disagree'' to ``agree'', using smiley faces~\cite{hall2016five}.  

\textcolor{black}{We also included two multiple choice questions asking children to justify which use-case was the most preferable and why, and which perception ability of the robots make them popular to the kids, as a means of getting more insight on their preferences. 
In general, the majority of children (12/31) stated that their favorite use-case was ``Pantomime'', due to the movements of the robot.}
\textcolor{black}{As we can see in Figure \ref{fig:subj}, the majority of the children (27/31) stated that they like playing with the robots, while 22 of them enjoyed playing because robots understood both their movements and speech. Many of them (21/31) also found the interaction and the use-cases easy to follow, without needing external help (19/31). Furthermore, children tended to agree (20 positive answers out of 31) that robots behave like humans. \textcolor{black}{By analyzing the questionnaire responses, we noticed that older children stated that they did not need prior knowledge in order to play with the robots, while younger children stated that they did.}}

\textcolor{black}{Similarly, in the assembly task that was evaluated in the primary school, 21 children were asked to express their opinion for the interaction. The questions are presented in Table~\ref{tab:ass_que}, and the available responses were a 3-point Likert scale (Disagree - Neutral - Agree). The Table also presents the questionnaire results after being mapped to a scale of 0-2, with 0 being the most negative. Their answers indicate that children were pleased with the interaction (1.81 MOS on whether they would like to play again with the robot and the comfortableness of the interaction). However, clearly the robot supervision for the assembly task has room for improvements, since although the instructions of the robot were very clear (1.95 MOS), children were neutral on whether they were actually helpful (1.10 MOS), or wrong (0.95 MOS).}

\textbf{Discussion}
\textcolor{black}{ In general, the evaluation of user experience during interaction with our multi-robot, multi-tasking, and multi-sensor robotic system provided encouraging results, proving that the system is technically capable of accommodating a complete CRI experience, with some adult intervention needed in certain cases and mainly for the collaborative task. Of course, there exists room for improvement, since many children stated that robots made mistakes frequently (18/31).
In the future, we aim to also conduct a subjective evaluation focused on the pedagogical aspect of the system, based on the insights collected during this initial study. }

\begin{table}[!h]
\centering
\scriptsize
\scalebox{1.2}{
\begin{tabular}{|c|c|} \hline
\multirow{2}{*}{\parbox{1cm}{Question}} & Mean Opinion \\
 & Score \\ \hline
Were you comfortable working with the robot? & 1.81 \\ \hline
Would you play with a robot again, sometime? & 1.81 \\ \hline
Was the robot helpful? & 1.10 \\ \hline
Did the robot make a lot of mistakes? & 0.95 \\ \hline
Were the robot’s instructions clear? & 1.95 \\ \hline
\end{tabular}}
\caption{Questions and results of the questionnaire presented to the children, following their ``Assembly Game'' with the robot.}
\label{tab:ass_que}
\end{table}



\section{Conclusion}
In this paper we have presented ChildBot, a multimodal perception framework that is the culmination and the extension of several earlier works by the authors in multimodal perception and CRI. ChildBot constitutes a CRI framework with multiple robotic agents that can be successfully used for edutainment purposes, and its perception system includes several different modules: audio-visual active speaker localization, and 6-DoF object tracking, visual activity recognition, and distant speech recognition. 
The effectiveness and successful interconnection of the modules has been demonstrated via \textcolor{black}{five indicative edutainment} CRI use-cases, each using a different subset of the various perception modules.

In order to validate the performance and the capabilities of our system for CRI, we have carried out an extensive objective evaluation of the developed perception modules, \textcolor{black}{as well as a user experience study that provides valuable initial insights for the interaction with ChildBot.} The experiments took place in a specially designed area that was decorated to resemble a child's room, where we collected both development data necessary for training the individual system modules, as well as use-case related data that were essential for testing the system performance during actual CRI.

Our results have showed that the individual perception technologies successfully capture the environment surrounding the interaction with a high degree of accuracy, \textcolor{black}{while the user experience study showed that children enjoyed playing with different robots.}

\textcolor{black}{For future work, we would like to also extend ChildBot for other applications, and it would be interesting to see how some of the novel perception methods we presented can generalize to other fields, such as rehabilitation or assistive applications for ASD children.}  
\textcolor{black}{Further, we aim to conduct a more thorough subjective evaluation on the pedagogical aspect of the system.}

In conclusion, our work shows that through the integration of multiple robots, sensors, and modalities, we can achieve a high level of unconstrained and autonomous CRI, opening up new prospects for the educational and entertainment social robotics of the future.
\section*{Acknowledgments}
This work has been funded by the BabyRobot project, supported by the EU Horizon 2020 under grant 687831.

The authors wish to thank Professor Christina Papaeliou and Dr. Asimenia Papoulidi for their help in designing the use-cases, supervising and evaluating the experiments with the children, and their  useful remarks.  We would like also to thank the ``Nea Genia Ziridi'' school for
their collaboration in the experimental procedure. Further, we wish to thank the BabyRobot project coordinators Professor Costas Tzafestas and Professor Alexandros Potamianos, Niko Kardari for providing his original code functions, Dr. Georgia Chalvatzaki for her help during the experiments, and the members of the NTUA IRAL. 

\newpage
\myclearpage

\bibliographystyle{plain}
\bibliographystyle{IEEEtranS}
\bibliography{bibaki.bib}

\end{document}